%% file: main.tex
\documentclass{article}

\PassOptionsToPackage{numbers, sort}{natbib}
\usepackage[main,final]{styles/neurips_2026}

\usepackage[utf8]{inputenc} % allow utf-8 input
\usepackage[T1]{fontenc}    % use 8-bit T1 fonts
\usepackage{hyperref}       % hyperlinks
\usepackage{url}            % simple URL typesetting
\usepackage{booktabs}       % professional-quality tables
\usepackage{amsfonts}       % blackboard math symbols
\usepackage{nicefrac}       % compact symbols for 1/2, etc.
\usepackage{microtype}      % microtypography
\usepackage{xcolor}         % colors
\usepackage{amsmath}
\usepackage{multirow}
\usepackage{graphicx}
\usepackage{wrapfig}
\usepackage{makecell}
\usepackage{subcaption}
\usepackage{cleveref}
\usepackage{longtable}
\usepackage{fontawesome5}
\usepackage[table]{xcolor}

\usepackage{siunitx}
\usepackage{xspace}
\newcommand{\method}{\textsc{ModelLens}\xspace}
\usepackage[most]{tcolorbox}
\usepackage{enumitem}
\definecolor{AppBlue}{HTML}{2F5C99}
\definecolor{AppLightBlue}{HTML}{F3F6FB}
\definecolor{AppGray}{HTML}{666666}
\newtcolorbox{appendixcard}[2]{
  enhanced,
  colback=AppLightBlue,
  colframe=AppBlue,
  boxrule=0.8pt,
  arc=2mm,
  left=2mm,
  right=2mm,
  top=1.5mm,
  bottom=1.5mm,
  title=\textbf{#1}\hfill{\small #2},
  coltitle=white,
  colbacktitle=AppBlue,
  fonttitle=\small,
  attach boxed title to top left={yshift=-1mm,xshift=2mm},
  boxed title style={arc=1mm,boxrule=0pt},
  before skip=6pt,
  after skip=6pt
}
\title{\textsc{ModelLens}: Finding the Best for Your Task \\
        from Myriads of Models}
\input{sections/author}

\begin{document}

\maketitle

\vspace{-3.0em}

\begin{center}
\faGithub\ Github:
\href{https://github.com/luisrui/ModelLens}
{https://github.com/luisrui/ModelLens.git}

\faGlobe\ Demo:
\href{https://huggingface.co/spaces/luisrui/ModelLens}
{huggingface.co/spaces/luisrui/ModelLens}
\end{center}

\input{sections/00abstract}

\input{sections/01intro}

\input{sections/02related_work}

\input{sections/04method}

\input{sections/05experiment}

\input{sections/06conclusion}

\section*{Acknowledgments}
We appreciate the reviewers for their constructive comments and suggestions. Rui Cai and Zhe Zhao were supported by NSF grant OAC 2531126 and a Google Research Award. Wenjie Jacky Mo, Xiaofei Wen and Muhao Chen were supported by a Noyce Initiative Grant, the NSF of the United States under Grants ITE 2333736 and OAC 2531126, the NIH 1R01CA323939-01, and equipment gifts from NVIDIA and DELL.

\newpage
{
\small
\bibliographystyle{unsrt}
\bibliography{sections/nips2026_conference}
}

\clearpage
\appendix
\input{sections/99appendix}

% \clearpage
% \input{sections/checklist}

\end{document}

%% file: sections/author.tex
\author{
Rui Cai$^{\dagger}$, Wenjie Jacky Mo$^{\dagger}$, Xiaofei Wen$^{\dagger}$, Qiyao Ma$^{\dagger}$, \\
{\bfseries Wenhui Zhu$^{\ddagger}$, Xiwen Chen$^{\S}$, Muhao Chen$^{\dagger}$, Zhe Zhao$^{\dagger}$} \\
$^\dagger$University of California, Davis \quad
$^\ddagger$Arizona State University \quad
$^\S$Morgan Stanley \\
\texttt{ruicai@ucdavis.edu}
}

%% file: sections/00abstract.tex
\begin{abstract}
The open-source model ecosystem now contains hundreds of thousands of pretrained models, yet picking the best model for a new dataset is increasingly infeasible: new models and unbenchmarked datasets emerge continuously, leaving practitioners with no prior records on either side.
Existing approaches handle only fragments of this \emph{in the wild} setting: AutoML and transferability estimation select models from small predefined pools or require expensive per-model forward passes on the target dataset, while model routing presupposes a given candidate pool.
We introduce \method, a unified framework for model recommendation in the wild. 
Our key insight is that public leaderboard interactions, though scattered and noisy, collectively trace out an implicit atlas of model capabilities across heterogeneous evaluation settings, a signal rich enough to learn from directly. By learning a performance-aware latent space over model--dataset--metric tuples, \method ranks unseen models on unseen datasets without running candidates on the target dataset.
On a new benchmark of 1.62M evaluation records spanning 47K models and 9.6K datasets, \method surpasses baselines that either rely on metadata alone or require running each candidate on the target dataset. Its recommended Top-K pools further improve multiple representative routing methods by up to 81\% relative EM gain across diverse QA benchmarks. Case studies on recently released benchmarks further confirm generalization to both text and vision-language tasks. 
\end{abstract}

%% file: sections/01intro.tex
\section{Introduction}

%  Problem, Background setting 
The rapid growth of open-source machine learning models has created an unprecedented opportunity for practitioners to build, customize, and deploy AI systems~\citep{paperwithcode, openllmleaderboard}. Platforms such as HuggingFace~\citep{huggingface} now host hundreds of thousands of models spanning diverse architectures, scales, and application domains. Faced with a new task or dataset, practitioners must decide which model to adopt or fine-tune for their specific use case. Despite its importance, this decision remains notoriously difficult, and typically demands extensive empirical evaluation or ad-hoc trial-and-error~\citep{automl, helm}. In this work, we take a step toward \textbf{model recommendation in the wild}, a setting in which thousands of heterogeneous models and datasets coexist across diverse architectures, modalities, and evaluation protocols.

\begin{figure}
\centering
    \includegraphics[width=\linewidth]{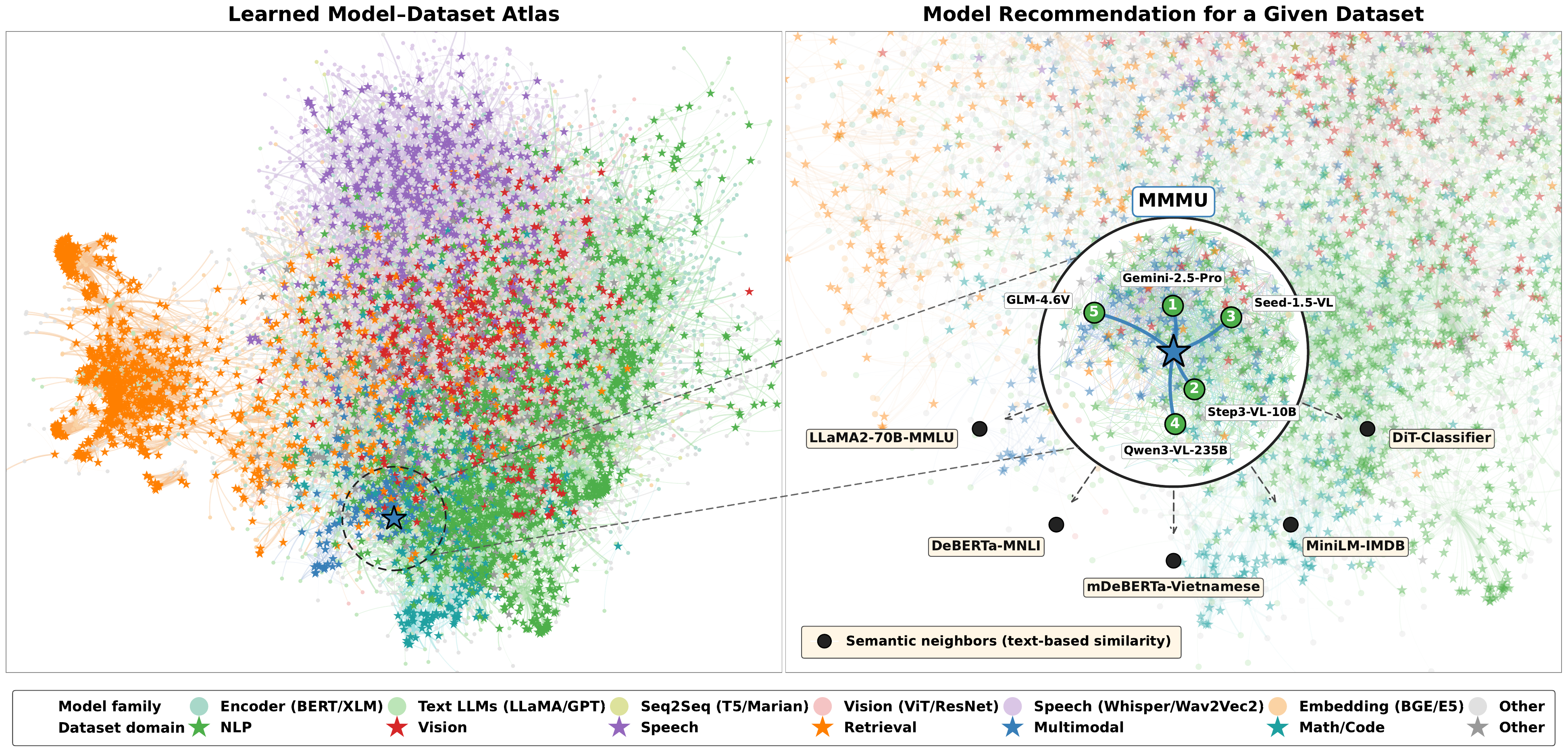}
    % \caption{Model recommendation in the wild. \textbf{(Left)} Atlas of $\sim$47K models (\textcolor{gray}{dots}) and $\sim$9.6K datasets (\textcolor{red}{$\star$}) laid out by a force-directed projection of our interaction-trained ecosystem structure rather than surface-level description similarity. The dashed circle marks the example dataset \textcolor[HTML]{377eb8}{\texttt{MMMU}}. \textbf{(Right)} Magnified view around \texttt{MMMU}: our framework retrieves the top-5 candidate models in this learned space (\textcolor[HTML]{4daf4a}{green numbered badges}). In contrast, nearest neighbours under \textit{raw} description embeddings (\textbf{black filled circles} reached by dashed     arrows) recover semantically related but performance-irrelevant models (e.g., DeBERTa-MNLI, DiT-Classifier) that lie far away in the learned atlas.}
    \caption{Model recommendation in the wild. \textbf{(Left)} Inspired by Model Atlas~\cite{modelatlas}, which maps the model ecosystem by lineage, we draw a 2D map of $\sim$47K models (\textcolor{gray}{dots}) and $\sim$9.6K datasets (\textcolor{red}{$\star$}), positioned by embeddings learned from model--dataset performance records. The dashed circle marks the example dataset \textcolor[HTML]{377eb8}{\texttt{MMMU}}. \textbf{(Right)} Zoom-in around \texttt{MMMU}: the top-5 models retrieved in this learned space (\textcolor[HTML]{4daf4a}{green numbered badges}) are competitive multimodal LLMs. In contrast, the nearest neighbours by description-embedding similarity alone (\textbf{black filled circles}, dashed arrows) are topically related but unsuitable for \texttt{MMMU} (e.g., DeBERTa-MNLI, DiT-Classifier), and lie far from it in the learned map.}
    \label{fig:teaser_figure_atlas}
    % \vspace{-0.6cm}
\end{figure}

% Limitations of Existing Work: No transferability, No Multimodal, Modern model constrains
However, existing approaches to model selection are ill-equipped for this in-the-wild setting.
Automated machine learning (AutoML) methods~\citep{zap, tabpfn, optimus} search over a fixed pool of models or pipelines to find the best fit for a target task. Transferability estimation~\citep{logme, modelspider, know2vec} ranks pretrained models for a given dataset, typically by extracting feature or label statistics from a forward pass on the target. Model routing~\citep{routerdc, embedllm, router-r1} performs instance-level selection over a predefined candidate pool, dispatching each query to one of a few pre-curated models. While each approach makes progress on a slice of the problem, none addresses the requirements of open model ecosystems along three axes.
\textit{Scale.} AutoML and routing presuppose a small, curated pool, ignoring the hundreds of thousands of models available today; Transferability estimation is pool-agnostic but requires a forward pass per candidate, infeasible at this scale.
\textit{Generalization.} Transferability and routing methods require evaluating each candidate on the target dataset, preventing extension to newly released models and unseen datasets.
% \textit{Heterogeneity.} All three lines of work assume homogeneous evaluation, with a single metric on a single task family, overlooking the metric incompatibility and task diversity of real-world benchmarks.
\textit{Heterogeneity.} All three lines of work assume homogeneous evaluation with a single metric on a single task family. Real benchmarks are heterogeneous even within a task family: captioning admits BLEU, ROUGE, CIDEr, and METEOR; classification admits accuracy, F1, and top-$k$ accuracy. These metrics can rank the same model differently, so single-metric conclusions are fragile.
These limitations raise a key question: 
\textit{can we leverage large-scale model–dataset interaction patterns to enable model selection in the wild, without requiring direct evaluation or fine-tuning?}
Our key insight is that the seemingly fragmented, large-scale interactions between models and datasets on modern leaderboards are not merely noise but a rich source of implicit supervision, encoding how model capabilities align with dataset characteristics.  
Figure~\ref{fig:teaser_figure_atlas} illustrates this on a real subset of our data: when models and datasets are projected into a space learned from interactions, they cluster naturally by modality and task type, whereas a space induced from textual descriptions alone fails to recover this structure (\Cref{fig:semantic_only_atlas}). For a target benchmark such as MMMU~\citep{mmmu}, the learned space surfaces the real competitive multimodal LLMs as nearest neighbors, while raw description similarity retrieves semantically related but performance-irrelevant models (e.g., DeBERTa-MNLI). 
This motivates formulating model recommendation as a learning problem over model–dataset interactions, providing recommendations without ever running candidate models on the target dataset. We instantiate this idea by aggregating performance records from public leaderboards~\citep{paperwithcode, openllmleaderboard, huggingface} into a unified repository, with each entry represented as a tuple (\textit{model, dataset, metric, performance}), and casting model recommendation in the wild as a ranking problem over these interactions.

Broadly, \method takes target dataset and candidate model descriptions together with leaderboard interactions as input, and outputs a ranking of candidates by predicted performance. Recommended models can be deployed via any downstream pipeline, such as zero-shot inference, in-context learning, fine-tuning, or routing. Specifically, \method introduces a structural prior over model scale and architecture family to capture predictable trends like neural scaling, paired with a learned interaction term for fine-grained model–dataset compatibility. To support cold-start inference on newly released models and unbenchmarked datasets, each entity is represented by identity, family, name, and description embeddings, with ID-dropout applied during training to force reliance on metadata when identity is unavailable.
We validate \method on 1.62M evaluation records spanning 47K models and 9.6K datasets, across matrix completion, held-out datasets, and newly released models. Despite leaderboard records mixing evaluation protocols (zero-shot, fine-tuning, prompting), the aggregated collaborative information proves useful: integrating \method's top-K outputs with modern routing methods yields gains of up to 81\% on QA benchmarks, and case studies on two recently released benchmarks confirm cross-modal transfer to text and vision-language tasks.

% Contributions

Our contributions are threefold:
1) We first formalize the problem of model recommendation in the wild and curate a large-scale benchmark of model–dataset–metric interactions, covering tens of thousands of models and diverse datasets across multiple domains and modalities.
2) We propose a unified, metric-aware ranking framework that leverages heterogeneous metadata to predict model-dataset compatibility, generalizing to unseen models and datasets without any direct evaluation or finetuning.
3)We show that the framework not only attains strong ranking performance, but also yields high-quality candidate sets directly compatible with downstream routing and ensemble systems, enabling scalable model selection in dynamic, large-scale ecosystems.

%% file: sections/02related_work.tex
\section{Related Works}

\textbf{Transferability Estimation.}
Transferability estimation predicts how well a pretrained model will transfer to a target task without full fine-tuning. Training-free methods estimate transferability from a single forward pass on the target dataset using information-theoretic or likelihood-based statistics~\citep{h-score, nce, leep, nleep, logme, pactran, otce, lfc, gbc}, while learning-based approaches model interactions between feature representations and target data~\citep{modelspider, know2vec}. Despite their effectiveness, TE methods assume a controlled pretrain-to-finetune pipeline and require per-model execution on the target dataset, which becomes infeasible as model hubs scale to tens of thousands of candidates~\citep{huggingface, openllmleaderboard}. \method instead studies model recommendation \emph{in the wild}: models are already fully specified systems, and rankings are predicted directly from large-scale leaderboard interactions and metadata, with forward-pass features supported as optional augmentations. A full taxonomy of TE (\Cref{apx:te_relateworks}).

\textbf{Automated Model Search.}
Automated machine learning (AutoML) aims to automate model selection and hyperparameter tuning for a target task. Classical approaches frame this as a search or meta-learning problem over a fixed pool of pipelines or architectures~\citep{automl, tabpfn}, with recent work extending this paradigm to pretrained model selection~\citep{zap, optimus}. While effective in curated settings, these methods assume a predefined and relatively small candidate pool, which fundamentally limits their applicability to the open and continuously evolving model ecosystems we target in this work.

\textbf{Model Routing.}
Model routing addresses an orthogonal problem: given a \emph{fixed} pool of candidates and an incoming query, decide which model should serve it~\citep{knnrouter, routellm, routerdc, embedllm, graphrouter, router-r1}. These methods take the candidate pool as given, leaving open the upstream question of how the pool itself should be constructed from a large, heterogeneous model space~\citep{routereval}. Our work is complementary: \method produces high-quality, task-specific candidate pools at the dataset level, which can be directly consumed by any instance-level router.

%% file: sections/04method.tex
\section{ModelLens}
\method is a ranking framework that predicts the relative performance of candidate models on a target dataset using heterogeneous metadata, without running any candidate on the target dataset. Its design follows a single principle: combine structured inductive bias with flexible interaction modeling. 
Three components instantiate this principle. First, \method builds \emph{multi-view representations} for models and datasets from learned IDs, tokenized names, and frozen text-description embeddings, supporting both memorization and generalization. Second, it conditions on the \emph{evaluation context} (task and metric) and on structural model attributes (scale and architecture family). Third, it computes compatibility via an additive decomposition into a \emph{structural prior} for predictable regularities such as neural scaling, and a \emph{residual interaction} term for fine-grained model--dataset compatibility. An \emph{ID dropout} mechanism applied during training enables zero-shot ranking on entirely new models or datasets. We first formalize the problem setting, then describe each component in turn.
% Concretely, \method (i) builds \emph{multi-view representations} for models and datasets from learned IDs, tokenized names, and frozen text-description embeddings, supporting both memorization of seen entities and generalization to unseen ones; (ii) conditions on the \emph{evaluation context} (task and metric) and on structural model attributes (scale and architecture family), which jointly govern performance across heterogeneous settings; and (iii) computes a compatibility score via an additive decomposition into a \emph{structural prior}, which captures predictable regularities such as neural scaling effects, and a \emph{residual interaction} term, which models fine-grained model--dataset compatibility beyond what structure alone explains. To prevent over-reliance on identity features and enable zero-shot ranking on entirely new models or datasets, we further introduce an \emph{ID dropout} mechanism during training. We first formalize the problem setting, then describe the feature representations, scoring function, and training objectives.

\input{sections/03preliminary}

\subsection{Feature Representation}
\label{sec:features}
% \method represents each model using a multi-view embedding that integrates both memorization-oriented and generalization-oriented signals.
\textbf{Model representation.}
Each model $m$ is encoded as the concatenation of three complementary parts:
\begin{equation}\label{eq:h-model}
  \mathbf{h}_m = [\,\mathbf{e}_m^{\mathrm{id}} \;\|\;
                    \mathbf{e}_m^{\mathrm{name}} \;\|\;
                    \mathbf{e}_m^{\mathrm{desc}}\,],
\end{equation}
where $\mathbf{e}_m^{\mathrm{id}} \in \mathbb{R}^{d_{\mathrm{id}}}$ is a learned ID embedding that captures model-specific behaviors observed during training, 
$\mathbf{e}_m^{\mathrm{name}} \in \mathbb{R}^{d_{\mathrm{tok}}}$ is a compositional name embedding obtained by tokenizing the model name and aggregating token embeddings, 
and $\mathbf{e}_m^{\mathrm{desc}} \in \mathbb{R}^{d_{\mathrm{desc}}}$ is a frozen semantic embedding of the model's textual description using a pretrained text encoder.
% This multi-view design separates memorization and generalization pathways: the learned components ($\mathbf{e}^{\mathrm{id}}$, $\mathbf{e}^{\mathrm{name}}$) capture fine-grained behaviors of seen models, while the frozen semantic embedding $\mathbf{e}^{\mathrm{desc}}$ provides a stable signal that remains informative for unseen models.

\textbf{Dataset representation.}
Each dataset $d$ is represented as:
\begin{equation}\label{eq:h-data}
  \mathbf{h}_d = [\,\mathbf{e}_d^{\mathrm{id}} \;\|\;
                    \mathbf{e}_d^{\mathrm{desc}}\,],
\end{equation}
where $\mathbf{e}_d^{\mathrm{id}} \in \mathbb{R}^{d_{\mathrm{ds\text{-}id}}}$ is a learned dataset ID embedding, and $\mathbf{e}_d^{\mathrm{desc}} \in \mathbb{R}^{d_{\mathrm{ds\text{-}desc}}}$ is a frozen semantic embedding of the dataset description from the same text encoder.
% The learned ID embedding captures dataset-specific patterns observed during training, while the frozen semantic embedding provides transferable information (e.g., domain and task characteristics) that remains informative for unseen datasets.

\textbf{Evaluation context and structural attributes.}
Beyond the model--dataset pair, performance also depends on \emph{how} a model is evaluated and on \emph{what kind} of model it is.
We encode the task type $t$ and metric $\mu$ as learned embeddings $\mathbf{e}_t \in \mathbb{R}^{d_{\mathrm{task}}}$ and $\mathbf{e}_\mu \in \mathbb{R}^{d_{\mathrm{metric}}}$, allowing the score to adapt to different evaluation protocols.
We further encode two structural attributes of each model: its scale,  discretized into size buckets and mapped to an embedding $\mathbf{e}_m^{\mathrm{size}} \in \mathbb{R}^{d_{\mathrm{size}}}$, capturing the non-linear and task-dependent effects of neural scaling; and its architecture family, represented by an embedding $\mathbf{e}_m^{\mathrm{fam}} \in \mathbb{R}^{d_{\mathrm{fam}}}$, encoding shared inductive biases among models derived from the same architecture.

% These structural attributes play a central role in our formulation: they serve as inputs to both the residual interaction model and the structural prior, enabling the model to explicitly reason about scaling laws and architecture-dependent performance patterns.
\subsection{Scoring Function: Residual + Prior Decomposition}
\label{sec:scoring_overview}

The compatibility score is decomposed additively into a \emph{structural prior} that depends only on model attributes and a \emph{residual interaction} that depends on the full evaluation context.
This separates two complementary sources of signal: predictable performance trends from model structure, and context-dependent affinity that cannot be explained by structure alone.

\textbf{Structural Prior.}
\label{sec:prior}
The structural prior $s_{\mathrm{prior}}(m)$ models the intrinsic competence of a model based solely on its structural attributes, independent of any specific dataset or task. It is parameterized as a shared function over model size and architecture family:
\begin{equation}\label{eq:prior}
  s_{\mathrm{prior}}(m)
  = \mathrm{MLP}_{\mathrm{prior}}\!\bigl(
      [\,\mathbf{e}_m^{\mathrm{size}} \;\|\;
         \mathbf{e}_m^{\mathrm{fam}}\,]
    \bigr) \;\in \mathbb{R}.
\end{equation}
This component explicitly models structural performance trends, such as neural scaling effects~\citep{kaplan2020scaling}, as a learnable function of model structure. Unlike per-model bias terms in collaborative filtering, $s_{\mathrm{prior}}$ is a shared parametric function over the (size, family) space, enabling generalization to unseen models by interpolating over this space.
By capturing predictable global patterns, the prior reduces the burden on the interaction model so that the residual can focus on fine-grained deviations.

\textbf{Residual Interaction.}
\label{sec:residual}
The residual term $s_{\mathrm{residual}}(m, d, t, \mu)$ models the deviation from the structural prior conditioned on the full evaluation context, capturing dataset-specific specialization and metric-dependent behavior. We concatenate all features into a joint input,
\begin{equation}\label{eq:residual-input}
  \mathbf{x} = [\,\mathbf{h}_m \;\|\;
                   \mathbf{h}_d \;\|\;
                   \mathbf{e}_m^{\mathrm{size}} \;\|\;
                   \mathbf{e}_m^{\mathrm{fam}} \;\|\;
                   \mathbf{e}_t \;\|\;
                   \mathbf{e}_\mu\,],
\end{equation}
which is passed through a multi-layer perceptron backbone to produce a hidden representation $\mathbf{h}$, followed by two linear heads:
\begin{equation}\label{eq:residual}
  \mathbf{h} = \mathrm{MLP}_{\mathrm{backbone}}(\mathbf{x}), \qquad
  s_{\mathrm{residual}} = \mathbf{w}_{\mathrm{pair}}^\top \mathbf{h}.
\end{equation}
The size and family embeddings are shared across both the prior and residual pathways: while the prior captures their \emph{marginal} effects, the residual captures \emph{interaction} effects, such as how the benefit of model scale varies across datasets or metrics.
In addition to the pairwise ranking score, the backbone also produces an auxiliary pointwise prediction:
\begin{equation}\label{eq:z-pred}
  \hat{z} = \mathbf{w}_{\mathrm{point}}^\top \mathbf{h},
\end{equation}
which estimates the standardized performance of a model on a dataset. This auxiliary objective encourages the shared representation to be informative for both ranking and regression.

\textbf{Score Composition.}
\label{sec:score-composition}
The final compatibility score combines the two components and rescales them by a learnable temperature:
\begin{equation}
  \tilde{s}(m, d, t, \mu)
  = \frac{s_{\mathrm{residual}}(m, d, t, \mu)
        + s_{\mathrm{prior}}(m)}
       {\max(\tau,\, \epsilon)}.
\end{equation}
The learnable temperature $\tau$ controls the sharpness of the result ranking distribution and $\epsilon$ is a constant to ensure numerical stability. The additive form also yields an interpretable decomposition: a model can be selected because of its general competence ($s_{\mathrm{prior}}$) and its task-specific affinity ($s_{\mathrm{residual}}$).

\subsection{Generalization via ID Dropout}
\label{sec:generalization}
Learned ID embeddings $\mathbf{e}_m^{\mathrm{id}}, \mathbf{e}_d^{\mathrm{id}}$ are powerful for memorization but useless for unseen entities at training time. To prevent the model from over-relying on them, during training we independently replace each ID embedding with a shared learnable $\textsc{[UNK]}$ vector with probabilities $p_m$ and $p_d$:
{\small
\begin{equation}\label{eq:id-dropout}
  \tilde{\mathbf{e}}_m^{\mathrm{id}} =
  \begin{cases}
    \mathbf{e}_{[\textsc{unk}]}^{\mathrm{model}} & \text{with probability } p_m, \\
    \mathbf{e}_m^{\mathrm{id}} & \text{otherwise},
  \end{cases}
  \qquad
  \tilde{\mathbf{e}}_d^{\mathrm{id}} =
  \begin{cases}
    \mathbf{e}_{[\textsc{unk}]}^{\mathrm{dataset}} & \text{with probability } p_d, \\
    \mathbf{e}_d^{\mathrm{id}} & \text{otherwise}.
  \end{cases}
\end{equation}
}

This trains a single set of parameters under two regimes simultaneously: a memorization regime when IDs are visible, and a semantic regime where the model must rely on names, descriptions, and structural attributes. At inference, unseen entities map to $\textsc{[UNK]}$ and are handled without any architectural change. 

\subsection{Multi-Objective Learning}
\label{sec:learning}
We supervise \method with three complementary objectives: pairwise comparisons capture local preferences, listwise likelihoods capture global ranking structure, and a pointwise regression captures absolute performance signals.

\textit{Pairwise ranking loss.}
Within each evaluation group, we sample pairs $(m^+, m^-)$ where $m^+$ outperforms $m^-$ and apply the BPR objective \cite{bpr}:
\begin{equation}\label{eq:bpr}
  \mathcal{L}_{\mathrm{pair}}
  = \mathbb{E}\!\left[\,
      -\log \sigma\!\bigl(\tilde{s}(m^+, d) - \tilde{s}(m^-, d)\bigr)
    \,\right].
\end{equation}

\textit{Listwise ranking loss.}
For each evaluation group with $M$ candidate models indexed in decreasing order of ground-truth performance, we adopt the Plackett--Luce likelihood \cite{plackett1975analysis,luce1959individual}:
\begin{equation}
\mathcal{L}_{\mathrm{list}} = \frac{1}{|\mathcal{G}|} \sum_{g \in \mathcal{G}} \frac{1}{M_g} \sum_{i=1}^{M_g} \left[ \log \sum_{j=i}^{M_g} \exp\!\left(\tilde{s}(m_j, d_g)\right) - \tilde{s}(m_i, d_g) \right].
\end{equation}

\textit{Pointwise regression loss.}
The auxiliary regression head is supervised against the standardized score $z(m, d)$, computed by z-scoring raw performance within each evaluation group; this within-group normalization is what makes scores comparable across heterogeneous metrics:
\begin{equation}\label{eq:point}
  \mathcal{L}_{\mathrm{point}}
  = \mathbb{E}\!\left[\,
      \bigl(\hat{z}(m, d) - z(m, d)\bigr)^2
    \,\right].
\end{equation}

\textbf{Final objective.}
The overall training objective is a weighted combination of the three losses:
\begin{equation}\label{eq:total-loss}
  \mathcal{L}
  = \lambda_{\mathrm{list}}\,\mathcal{L}_{\mathrm{list}}
  + \lambda_{\mathrm{pair}}\,\mathcal{L}_{\mathrm{pair}}
  + \lambda_{\mathrm{point}}\,\mathcal{L}_{\mathrm{point}}.
\end{equation}
The pairwise and listwise losses operate on $\tilde{s}$ and jointly train both the prior and residual pathways, while the pointwise loss grounds the shared backbone in absolute performance magnitudes.

%% file: sections/03preliminary.tex
% \subsection{Preliminary}
% \label{sec:preliminary}
\subsection{Problem Definition}

Let $\mathcal{M} = \{m_1, \dots, m_N\}$ denote a large and evolving pool of available models, and $\mathcal{D} = \{d_1, \dots, d_T\}$ a collection of datasets. 
% Each $m_i$ represents a publicly available model checkpoint, which may correspond to a pretrained, instruction-tuned, or fine-tuned variant. 
Each pair $(m_i, d_j)$ is associated with a performance score $y_{ij} \in \mathbb{R}$ under a task-specific evaluation metric, forming a performance matrix $\mathbf{Y} \in \mathbb{R}^{N \times T}$ whose observed entries are
\begin{equation}
\mathcal{O} = \{(m_i, d_j, y_{ij}) \mid (i,j) \in \Omega \}.
\end{equation}
In practice, $\mathbf{Y}$ is sparse and heterogeneous: few pairs are evaluated, metrics differ across datasets so absolute scores are not directly comparable.
Given a target dataset $d^\ast$ with limited or no observed evaluations, the goal of \emph{model recommendation in the wild} is to learn a scoring function
\begin{equation}
f: \mathcal{M} \times \mathcal{D} \times \mathcal{T} \times \mathcal{U} \rightarrow \mathbb{R},
\end{equation}
where $\mathcal{T}$ and $\mathcal{U}$ are the spaces of task types and evaluation metrics (necessary since the same pair can rank differently under different metrics, e.g., accuracy vs.\ F1). For a target dataset $d^\ast$ evaluated under metric $\mu^\ast$ and task $t^\ast$, the framework produces:
\begin{equation}
m^\ast = \arg\max_{m \in \mathcal{M}} f(m, d^\ast, t^\ast, \mu^\ast), \qquad
\mathcal{M}_K = \operatorname{TopK}_{m \in \mathcal{M}} f(m, d^\ast, t^\ast, \mu^\ast).
\end{equation}
Crucially, $f$ takes only model and dataset descriptors together with the evaluation context $(t,\mu)$ as input, and does \emph{not} consume any feature, gradient, or forward-pass signal extracted from $d^\ast$. 
Since metrics are incompatible across datasets, we supervise $f$ via the \emph{relative ordering} of models within each evaluation group $g = (d, t, \mu) \in \mathcal{G}$, where $\mathcal{G}$ denotes the set of all (dataset, task, metric) groups observed in training rather than their absolute values, and the central challenge is to generalize this ranking to unseen models and datasets under sparse, heterogeneous observations $\mathcal{O}$.

%% file: sections/05experiment.tex
\section{Experiments}
In the experiments, we aim to answer the following questions:

Q1: Can our method accurately model model--dataset interactions, both in terms of recovering missing entries and generalizing to unseen datasets and models?

Q2: How does our method perform under standard transferability-based model selection settings?

Q3: Can dataset-level model recommendation improve instance-level routing?

\subsection{Dataset Construction}
We construct a large-scale dataset for \emph{Model Recommendation in the Wild}, where the goal is to rank candidate models for a given dataset without direct evaluation. 
Unlike prior work focused on small or single-domain settings, our dataset captures heterogeneous model--dataset interactions across diverse tasks and modalities.
We aggregate records from three public sources: HuggingFace Model Hub~\citep{huggingface}, Open LLM Leaderboard~\citep{openllmleaderboard}, and PapersWithCode~\citep{paperwithcode}, with HuggingFace records extracted via a three-tier pipeline prioritizing structured YAML, model-card metadata, and LLM-parsed README tables in decreasing reliability~\citep{modeltables}.
After deduplication, the dataset contains 1.62M records over 47K models and 9.6K datasets, spanning 2{,}551 tasks, and 348 architecture families across multiple domains.
To evaluate generalization, we support two complementary settings: \emph{performance completion}, where masked entries from observed datasets are predicted, and \emph{cold-start generalization}, where 609 datasets and 375 models (temporally partitioned due to public released timestamps) are held out entirely from training. 
% For unseen-model evaluation, held-out models are split by public release time, simulating recommendation for newly released models without prior interaction observations.
Dataset and model splits are further stratified across task type and modality to reduce domain skew.
Full details are in~\Cref{sec:dataset}.
\subsection{Model Recommendation in the Wild}

\textbf{Baselines and Evaluation Metrics.} 
We compare against model selection methods from two paradigms, depending on whether they require running candidates on the target dataset. 
\textit{Feature-based transferability methods} compute per-model scores from a forward pass on the target dataset, including training-free metrics (\texttt{H-Score}~\citep{h-score}, \texttt{NCE}~\citep{nce}, \texttt{LEEP}~\citep{leep}, \texttt{NLEEP}~\citep{nleep}, \texttt{LogME}~\citep{logme}, \texttt{PACTran}~\citep{pactran}, \texttt{OTCE}~\citep{otce}, \texttt{LFC}~\citep{lfc}, \texttt{GBC}~\citep{gbc}) and learning-based meta-rankers (\texttt{Model-Spider}~\cite{modelspider}, \texttt{Know2Vec}~\citep{know2vec}).
\textit{Feature-free methods} rely on metadata or learned interactions: \texttt{Task2Vec}~\citep{task2vec}, \texttt{ZAP}~\citep{zap}, and two practitioner-heuristic strawmen, \texttt{Model Size} (parameter count) and \texttt{Model Popularity} (HuggingFace downloads). Details  in~\Cref{sec:baseline_details}.
We evaluate ranking quality using Kendall's weighted \textit{$\tau_{w}$}~\citep{kendalltau} as the primary metric, which emphasizes top-rank correctness, and further report \textit{Hit@$K$}, \textit{NDCG@$K$}, and \textit{Rec@$K$}, averaged per dataset across the test set.

\begin{table*}[t]
\centering
\caption{Model ranking performance under new model and new dataset evaluation settings. Best results are in bold, and second-best results are underlined.}
\label{tab:generalization}
\setlength{\tabcolsep}{1.5pt}
\scriptsize
%\resizebox{\textwidth}{!}{%
\begin{tabular}{>{\raggedright\arraybackslash}llc|ccc|ccc|ccc}
\toprule
\textbf{Setting} & \textbf{Method} & \textbf{$\tau_{w}$} 
& \textbf{NDCG@1} & \textbf{Hit@1} & \textbf{Rec@1}
& \textbf{NDCG@10} & \textbf{Hit@10} & \textbf{Rec@10}
& \textbf{NDCG@30} & \textbf{Hit@30} & \textbf{Rec@30} \\
\midrule
\multirow{4}{*}{\makecell[l]{\textbf{Performance}\\ \textbf{Completion} \\ (2967 datasets)}}
& \textbf{ModelLens} 
& \textbf{0.868} 
& \textbf{0.954} & \textbf{0.153} & \textbf{0.153} 
& \textbf{0.967} & \textbf{0.521} & \textbf{0.452} 
& \textbf{0.974} & \textbf{0.840} & \textbf{0.764} \\
& ZAP
& \underline{0.763} 
& \underline{0.903} & 0.115 & 0.115 
& \underline{0.922} & \underline{0.517} & 0.369 
& \underline{0.937} & \underline{0.807} & \underline{0.642} \\
& Task2Vec 
& 0.417 
& 0.847 & \underline{0.132} & \underline{0.132} 
& 0.869 & 0.361 & \underline{0.315} 
& 0.884 & 0.646 & 0.500 \\
& Model Size 
& -0.021 
& 0.625 & 0.032 & 0.032 
& 0.716 & 0.129 & 0.167 
& 0.775 & 0.415 & 0.399 \\
& Model Popularity  
& -0.035
& 0.716 & 0.016 & 0.016 
& 0.704 & 0.078 & 0.071 
& 0.724 & 0.213 & 0.212 \\
\midrule
\multirow{4}{*}{\shortstack{\textbf{New Datasets} \\ (609 datasets)}}
& \textbf{ModelLens}
& \textbf{0.745} 
& \textbf{0.910} & \textbf{0.266} & \textbf{0.266} 
& \textbf{0.951} & \textbf{0.456} & \textbf{0.303} 
& \textbf{0.962} & \textbf{0.666} & \textbf{0.631} \\
& ZAP 
& \underline{0.253} 
& \underline{0.852} & \underline{0.060} & \underline{0.060} 
& \underline{0.861} & 0.189 & \underline{0.270} 
& \underline{0.870} & \underline{0.543} & \underline{0.482} \\
& Task2Vec 
& 0.227 
& 0.691 & 0.008 & 0.008 
& 0.778 & \underline{0.221} & 0.129 
& 0.817 & 0.365 & 0.381 \\
& Model Size  
& 0.059 
& 0.621 & 0.036 & 0.036  
& 0.721 & 0.117 & 0.190 
& 0.780 & 0.430 & 0.421 \\
& Model Popularity   
& -0.104
& 0.744 & 0.017 & 0.017 
& 0.707 & 0.072 & 0.058 
& 0.720 & 0.183 & 0.205 \\
\midrule
\multirow{4}{*}{\shortstack{\textbf{New Models} \\ (375 models)}}
& \textbf{ModelLens} 
& \textbf{0.402} 
& \textbf{0.929} & \textbf{0.009} & \textbf{0.009} 
& \textbf{0.923} & \textbf{0.137} & \textbf{0.210} 
& \textbf{0.932} & \textbf{0.412} & \textbf{0.480} \\
& ZAP 
& \underline{0.307} 
& \underline{0.884} & 0.004 & 0.004 
& \underline{0.913} & 0.072 & \underline{0.165} 
& \underline{0.920} & 0.299 & \underline{0.469} \\
& Task2Vec
& 0.078 
& 0.844 & 0.000 & 0.000 
& 0.870 & \underline{0.109} & 0.139 
& 0.879 & 0.310 & 0.250 \\
& Model Size  
& 0.055 
& 0.674 & \underline{0.005} & \underline{0.005} 
& 0.807 & 0.086 & 0.097 
& 0.853 & \underline{0.347} & 0.464 \\
& Model Popularity   
& -0.296
& 0.861 & 0.004 & 0.004 
& 0.858 & 0.088 & 0.109 
& 0.839 & 0.262 & 0.251 \\
\bottomrule
\end{tabular}
%}
% \vspace{-0.4cm}
\end{table*}

\subsubsection{Performance Completion and Cold-start Generalization (Q1)}

\textbf{Setup.}
We evaluate our method under two complementary settings:
(1)\emph{Performance Completion.}
From a partially observed performance matrix over 2,967 datasets, we randomly mask a subset of observed entries and train the model to predict their values, then derive a full ranking over candidate models from the predicted scores. This setting evaluates whether the model can recover global interaction structure from incomplete observations.
(2)\emph{Cold-start Generalization.}
We further evaluate two extrapolation scenarios: \emph{Unseen datasets} and \emph{Unseen models}, each requiring the model to generalize beyond observed interactions.
At this scale, feature-based transferability estimation methods are computationally infeasible since they require a forward pass per candidate on each target dataset; we therefore restrict comparison to feature-free baselines.

% For both settings, we evaluate ranking quality using Kendall’s $\tau$ and top-$K$ metrics, including NDCG@$K$, Hit@$K$, and Rec@$K$.
% All metrics are computed per dataset based on the full predicted ranking over candidate models.

\textbf{Results.}
Table~\ref{tab:generalization} shows that \method consistently outperforms all baselines across both performance completion and cold-start settings. The advantage is most pronounced on unseen datasets, where baselines degrade sharply while \method remains strong, indicating that the learned representations transfer beyond observed pairs to entirely new datasets and models. A per-task-family breakdown, including where \method falls short, is given in Appendix~\ref{app:task_family}.

\begin{table*}[t]
\caption{Transferability-based model selection on 8 vision benchmarks, measured by $\tau_{w}$. }
\label{tab:tau}
\centering
\scriptsize
\setlength{\tabcolsep}{5pt}
%\resizebox{\textwidth}{!}{%
\begin{tabular}{lccccccccc}
\toprule
\textbf{Method}
& \textbf{Aircraft} & \textbf{Cars} & \textbf{DTD} & \textbf{Pets} & \textbf{Flowers102} & \textbf{Food101} & \textbf{Country211} & \textbf{EuroSAT} & \textbf{Avg.} \\
\midrule

\multicolumn{10}{c}{\textbf{\textit{Feature-based Transferability Methods}}} \\
\midrule
H-Score~\citep{h-score}      & 0.328 & 0.616 & 0.395 & 0.610 & -0.200 & 0.200 & -0.629 & -0.067 & 0.157 \\
NCE~\citep{nce}          & 0.501 & 0.771 & 0.403 & 0.696 & -0.200 & -0.378 & \bfseries 0.511 & -0.200 & 0.263 \\
LEEP~\citep{leep}         & 0.244 & 0.704 & -0.111 & 0.680 & -0.111 & -0.022 & 0.074 & -0.244 & 0.152 \\
NLEEP~\citep{nleep}        & -0.725 & 0.622 & 0.074 & 0.787 & 0.244 & -0.378 & -0.422 & -0.156 & 0.005 \\
LogME~\citep{logme}        & 0.540 & 0.677 & 0.429 & 0.628 & -0.511 & 0.067 & 0.422 & -0.422 & 0.229 \\
PACTran~\citep{pactran}      & 0.031 & 0.665 & -0.236 & 0.616 & 0.022 & -0.067 & -0.270 & 0.067 & 0.104 \\
OTCE~\citep{otce}         & -0.241 & -0.157 & -0.165 & 0.402 & -0.111 & -0.289 & -0.405 & 0.333 & -0.079 \\
LFC~\citep{lfc}          & 0.279 & 0.243 & -0.722 & 0.215 & 0.244 & 0.467 & 0.405 & 0.478 & 0.201 \\
GBC~\citep{gbc}          & -0.744 & -0.265 & -0.102 & 0.163 & 0.289 & -0.022 & 0.384 & -0.200 & -0.062 \\
\midrule
Model-Spider~\citep{modelspider} & 0.467 & 0.644 & 0.556 & 0.689 & -0.556 & 0.067 & 0.244 & 0.289 & 0.300 \\
Know2Vec~\citep{know2vec}     & 0.111 & 0.283 & 0.200 & 0.200 & 0.067 & -0.156 & 0.289 & 0.244 & 0.155 \\

\midrule
\multicolumn{10}{c}{\textbf{\textit{Feature-free Methods}}} \\
\midrule
Task2Vec~\citep{task2vec}  & 0.272 & 0.404 & -0.279 & 0.426 & -0.263 & -0.511 & -0.422 & 0.460 & 0.011 \\
ZAP~\citep{zap}           & 0.244 & 0.188 & 0.244 & 0.246 & 0.067 & 0.378 & 0.315 & 0.156 & 0.229 \\
GPT-5~\citep{gpt5} & 0.142 & 0.249 & 0.002 & 0.727 & -0.347 & \bfseries 0.602 & 0.355 & -0.125 & 0.201 \\

\midrule
\multicolumn{10}{c}{\textbf{\textit{Ours}}} \\
\midrule
ModelLens (Feature Free) & 0.378 & 0.556 & 0.289 & 0.511 & 0.156 & 0.422 & 0.378 & 0.263 & 0.369 \\
\textbf{ModelLens (Feature Aug.)}
& \bfseries 0.556 & \bfseries 0.778 & \bfseries 0.689 &  \bfseries 0.802
& \bfseries 0.422 &  0.556 &  0.467 &  \bfseries 0.600
& \bfseries 0.609 \\

\bottomrule
\end{tabular}%
%}
\vspace{-0.4cm}
\end{table*}

\subsubsection{Transferability-based Model Selection (Q2)}
\textbf{Setups.} 
% For completeness, we also evaluate \method under the standard transferability-based model selection protocol~\citep{logme}, on 8 widely-used vision benchmarks: Aircraft~\citep{aircraft}, Cars~\citep{cars}, DTD~\citep{dtd}, Pets~\citep{pets}, Flowers102~\citep{flowers102}, Food101~\citep{food101}, Country211~\citep{country211}, and EuroSAT~\citep{eurosat}. 
For completeness, we also evaluate \method under the standard transferability-based model selection protocol~\citep{logme}, on 8 widely-used vision benchmarks: Aircraft~\citep{aircraft}, Cars~\citep{cars}, DTD~\citep{dtd}, Pets~\citep{pets}, Flowers102~\citep{flowers102}, Food101~\citep{food101}, Country211~\citep{country211}, and EuroSAT~\citep{eurosat}. The first four are in-distribution for learning-based meta rankers, while the latter four are unseen, allowing us to probe both regimes. We use GPT-5~\citep{gpt5} for direct recommendation, serving as another feature-free baseline.
Ranking quality is measured by Kendall’s weighted $\tau$, with MRR results in~\Cref{tab:mrr}. 

\textbf{Feature augmentation with transferability signals and Results.}
We further investigate whether forward-pass features from candidate models provide complementary information.
We extract intermediate representations from candidate models on the target dataset (as in feature-based baselines) and concatenate them with our existing model representations.
To prevent leakage, only auxiliary models that are disjoint from the evaluation pool contribute forward-pass features at training time, while features from evaluated candidates are used exclusively at inference.
Table~\ref{tab:tau} reports per-dataset and average $\tau_{w}$
on the 8 benchmarks. 
Our method attains the best average performance among all baselines without any forward pass on the target dataset. Feature-based transferability estimators are unstable across datasets rather than uniformly weaker (e.g., NLEEP ranges from $-0.725$ on Aircraft to $0.787$ on Pets). Adding transferability features (Feature Aug.) yields complementary gains: the average $\tau_w$ rises to 0.609, the best on six of eight datasets and on average.

\subsection{Routing with Recommended Model Pools}

\begin{wraptable}{r}{0.49\textwidth}
\vspace{-1.2cm}
\centering
\scriptsize
\setlength{\tabcolsep}{2.5pt}
\renewcommand{\arraystretch}{0.95}
\caption{Model pool replacement for NQ under comparable inference scale (recommended pools for other datasets in~\Cref{apx:pools}).}
\label{tab:model_replace}
\begin{tabular}{l c c c l}
\toprule
\textbf{Original} & \textbf{Scale} & $\rightarrow$ & \textbf{Scale} & \textbf{Selected} \\
\midrule
LLaMA-3.1-70B~\citep{llama31}   & $\approx 70$B & $\rightarrow$ & $\approx 70$B & LLaMA-3.3-70B~\citep{llama33} \\
Mixtral-8x22B~\citep{mixtral}   & $\approx 39$B & $\rightarrow$ & $\approx 20$B & GPT-OSS-20B~\citep{gptoss20b} \\
Gemma-2-27B~\citep{gemma}     & $\approx 27$B & $\rightarrow$ & $\approx 17$B & Llama-4-Maverick~\citep{llama4} \\
LLaMA-3.1-8B~\citep{llama31}    & $\approx 8$B  & $\rightarrow$ & $\approx 8$B  & Nemotron-H-8B-R~\citep{nemotron-h}\\
Qwen2.5-7B~\citep{qwen2.5}      & $\approx 7$B  & $\rightarrow$ & $\approx 7$B  & Qwen2.5-7B~\citep{qwen2.5} \\
Mistral-7B~\citep{mistral}      & $\approx 7$B  & $\rightarrow$ & $\approx 4$B  & Gemma-3n-E4B~\citep{gemma} \\
\bottomrule
\end{tabular}
\vspace{-0.5cm}
\end{wraptable}

\textbf{Setups.}
We evaluate instance-level model routing on five QA benchmarks: NQ~\citep{nq}, PopQA~\citep{popqa}, HotpotQA~\citep{hotpotqa}, Musique~\citep{musique}, and Bamboogle~\citep{bamboogle}, following the setup of Router-R1~\citep{router-r1}.
Unlike prior work that improves routing algorithms under a fixed model pool, we study the impact of \emph{pool quality} on routing performance. 
We evaluate multiple routing methods, including KNNRouter~\citep{knnrouter}, MLPRouter~\citep{routellm}, RouterDC~\citep{routerdc}, GraphRouter~\citep{graphrouter}, and Router-R1~\citep{router-r1}, evaluated with both the original pool and  recommended pool.

\begin{table*}[t]
\centering
\scriptsize
\setlength{\tabcolsep}{3pt}
\caption{Routing performance w.r.t Exact Match. 
Each method is evaluated with its original pool and with new model pool (\textit{Recommended Pool}).}
\label{tab:routing}
\begin{tabular*}{\textwidth}{@{\extracolsep{\fill}}lcccccc}
\toprule
\textbf{Method} & \textbf{NQ} & \textbf{PopQA} & \textbf{HotpotQA} & \textbf{Musique} & \textbf{Bamboogle} & \textbf{Avg.} \\
\midrule
KNNRouter~\citep{knnrouter}      & 0.262 & 0.222 & 0.224 & 0.066 & 0.360 & 0.227 \\
\quad + Recommended Pool    & \textbf{0.487 (↑85.9\%)} & \textbf{0.537 (↑141.9\%)} & \textbf{0.330 (↑47.3\%)} & \textbf{0.101 (↑53.0\%)} & \textbf{0.600 (↑66.7\%)} & \textbf{0.411 (↑81.1\%)} \\
\midrule
MLPRouter~\citep{routellm}     & 0.252 & 0.222 & 0.198 & 0.072 & 0.360 & 0.221 \\
\quad + Recommended Pool    & \textbf{0.475 (↑88.5\%)} & \textbf{0.490 (↑120.7\%)} & \textbf{0.251 (↑26.8\%)} & \textbf{0.096 (↑33.3\%)} & \textbf{0.520 (↑44.4\%)} & \textbf{0.366 (↑65.6\%)} \\
\midrule
RouterDC~\citep{routerdc}       & 0.278 & 0.282 & 0.244 & 0.080 & 0.504 & 0.278 \\
\quad + Recommended Pool      & \textbf{0.325 (↑16.9\%)} & \textbf{0.389 (↑37.9\%)} & \textbf{0.350 (↑43.4\%)} & \textbf{0.115 (↑43.8\%)} & \textbf{0.512 (↑1.6\%)} & \textbf{0.338 (↑21.6\%)} \\
\midrule
GraphRouter~\citep{graphrouter}    & 0.276 & 0.280 & 0.234 & 0.076 & 0.448 & 0.263 \\
\quad + Recommended Pool   & \textbf{0.405 (↑46.7\%)} & \textbf{0.600 (↑114.3\%)} & \textbf{0.264 (↑12.8\%)} & \textbf{0.132 (↑73.6\%)} & \textbf{0.584 (↑30.4\%)} & \textbf{0.397 (↑51.0\%)} \\
\midrule
% KNNRouter$^{*}$~\citep{knnrouter}       & 0.236 & 0.232 & 0.154 & 0.072 & 0.384 & 0.216 \\
% \quad + Recommended Pool   & \textbf{0.407 (↑72.5\%)} & \textbf{0.419 (↑80.6\%)} & \textbf{0.232 (↑50.6\%)} & \textbf{0.098 (↑36.1\%)} & \textbf{0.520 (↑35.4\%)} & \textbf{0.335 (↑55.1\%)} \\
% \midrule
Router-R1-Qwen~\citep{router-r1}      & 0.388 & 0.384 & 0.352 & 0.138 & 0.512 & 0.355 \\
\quad + Recommended Pool     & \textbf{0.524 (↑35.1\%)} & \textbf{0.501 (↑30.5\%)} & \textbf{0.538 (↑52.8\%)} & \textbf{0.224 (↑62.3\%)} & \textbf{0.624 (↑21.9\%)} & \textbf{0.482 (↑35.8\%)} \\
\bottomrule
\end{tabular*}
% \vspace{-0.2cm}
\end{table*}

\textbf{Model Pool Construction.}
We start from the six-model pool released with Router-R1~\citep{router-r1} (the \emph{original pool}). For each test dataset (held out from training), \method predicts model rankings from the dataset's textual description and evaluation metric alone, without access to any ground-truth performance. 
We then replace each model in the original pool with a top-ranked alternative of comparable scale that is available via the Together AI API\footnote{\url{https://www.together.ai/}} and matched on inference cost (parameter count for dense models, active parameters for MoE), ensuring both competitive ranking quality and deployability. Table~\ref{tab:model_replace} illustrates the resulting NQ pool; comparisons against recency-based and ZAP-selected pools are in~\Cref{app:pool_controls}. The procedure is orthogonal to the underlying router and can be applied to any existing method.

\textbf{Results.}
Table~\ref{tab:routing} shows routing performance under different model pools. Replacing the original pool with our recommended pool consistently improves all five routers across all five datasets, indicating that pool quality is orthogonal to and complementary with routing algorithm design.
% Table~\ref{tab:routing} shows routing performance with different model pools.
% Across all routing methods, replacing the original model pool with our selected pool consistently improves performance. 
% The gains are substantial across datasets, demonstrating that better model selection directly translates to improved routing performance.
% This trend holds across diverse routing strategies, indicating that our approach provides a stronger foundation for downstream routing.

\subsection{Ablation Study}
\textbf{Loss ablation and Results.}
We ablate the three training objectives: listwise (L), pairwise (P), and pointwise regression (Pt). 
The full model (L+P+Pt) achieves the best $\tau_w$ of 0.745. 
Removing the listwise loss causes the largest degradation ($\to$ 0.632), confirming that global ranking structure is the dominant supervision signal; removing pairwise yields a moderate drop ($\to$ 0.703), and removing pointwise the smallest ($\to$ 0.728), indicating that it primarily serves as calibration. 
Single-loss variants underperform almost all multi-loss combinations, showing the three signals are complementary. 
Full results are in~\Cref{tab:ablation_loss}. Further analyses of model-side and dataset-side feature contributions and unseen-family generalization are provided in~\Cref{apx:feature_ablation,app:family-holdout}.

\begin{figure}[t]
    \centering
    \begin{subfigure}{0.48\textwidth}
        \centering
        \includegraphics[width=\linewidth]{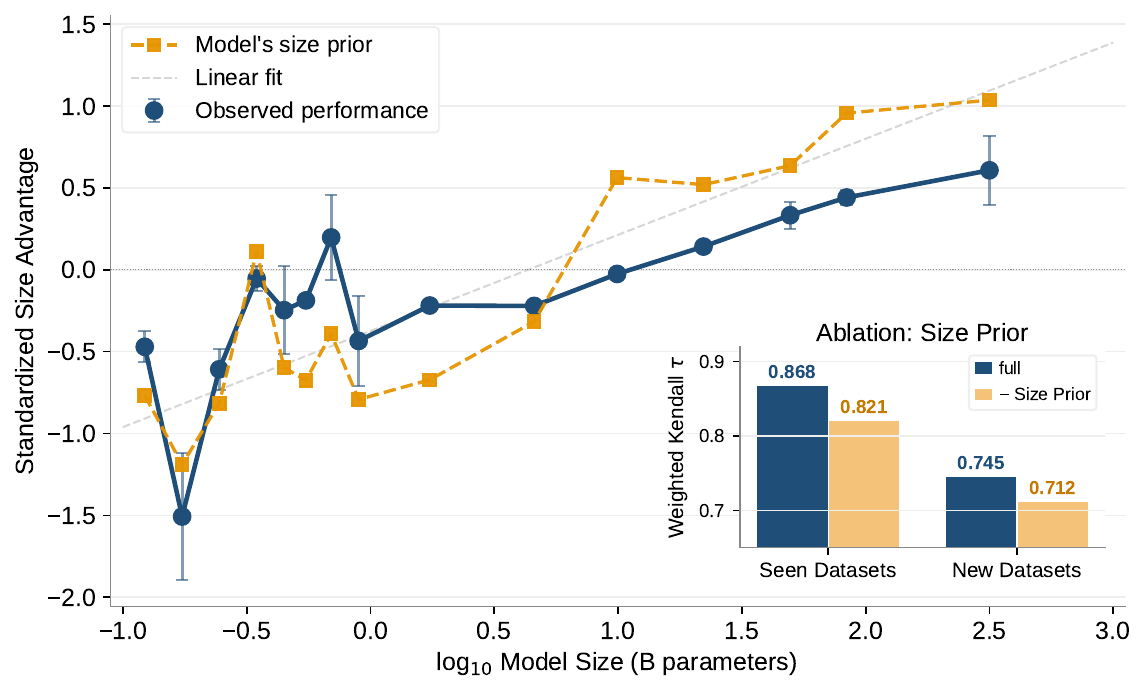}
    \end{subfigure}
    \hfill
    \begin{subfigure}{0.48\textwidth}
        \centering
        \includegraphics[width=\linewidth]{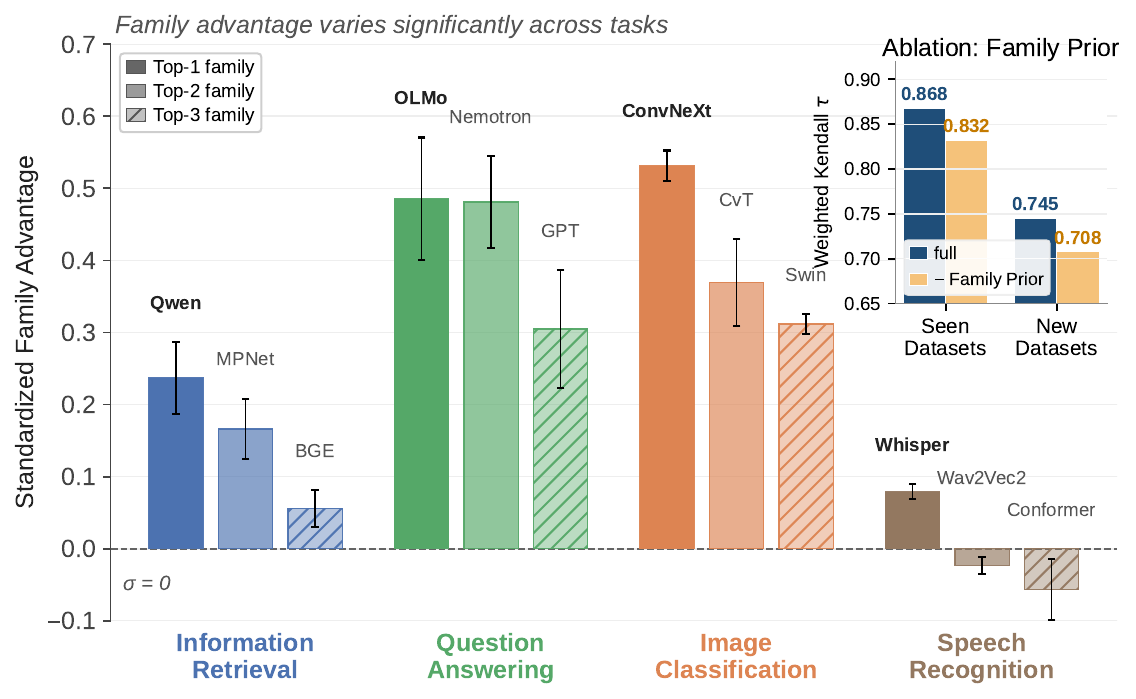}
    \end{subfigure}
    % \vspace{-4pt}
    \caption{Learned size and family priors from model–dataset interactions. (Left) Model performance exhibits a monotonic trend with respect to model size, with higher variance in the small-model regime. (Right) Family-level advantages vary across task domains, showing strong effects in some tasks (e.g., QA, IR, vision) but weaker structure in others (e.g., speech). Ablations confirm that both priors contribute to model recommendation performance.}
    % \caption{Learned size and family priors. (Left) Performance scales monotonically with model size; the small-model regime ($<1$B) is noisier. (Right) Family advantage is strong in QA, IR, and vision, weak in speech. Insets ablate each prior.}
    % \caption{Learned size and family priors. (Left) Performance scales monotonically with model size; (Right) Family advantage is strong in QA, IR, and vision, weak in speech. Insets ablate each prior.}
    \label{fig:prior}
    % \vspace{-0.4cm}
\end{figure}

% \begin{figure}[t]
%     \centering
%     \begin{subfigure}{0.48\textwidth}
%         \centering
%         \includegraphics[width=\linewidth]{figs/ngqa_task_adaptive_recommendation.pdf}
%     \end{subfigure}
%     \hfill
%     \begin{subfigure}{0.48\textwidth}
%         \centering
%         \includegraphics[width=\linewidth]{figs/rsvlm_qa_ranker_vs_meteor.pdf}
%     \end{subfigure}
%     \vspace{-4pt}
%     \caption{Case studies on unseen datasets across domains. (Left) On NGQA, different tasks favor different models, and the recommendations consistently outperform the default baseline. (Right) On RSVLM-QA, predicted scores perfectly match empirical performance, recovering the full ranking and generalizing to additional competitive models not included in the original benchmark.}
%     \label{fig:case_study}
% \end{figure}

\textbf{Learned Size and Family Priors.}
We analyze whether \method captures structured patterns from interactions, focusing on size and family effects. To enable comparison across heterogeneous tasks and metrics, we standardize performance via z-scores within each evaluation group and report group-level averages as size or family advantage (\Cref{apx:prior}).
\textit{Size prior.}
Figure~\ref{fig:prior}(left) shows a monotonic relationship between model size and predicted performance, aligning with empirical scaling trends. 
The effect is less stable for small models ($<1$B), where the regime is dominated by specialized models (e.g., vision models on vision tasks).
\textit{Family prior.}
Figure~\ref{fig:prior}(right) shows strong family-level effects in information retrieval, question answering, and image classification, where certain families consistently dominate; the effect is weaker in speech, where families perform more uniformly.
% These observations suggest that different priors operate under different regimes: size provides a global trend that becomes reliable for sufficiently large models, while family captures task-dependent structure that is prominent in some domains but weak in others.
\textit{Impact on recommendation.}
The two priors are complementary: size provides a global trend that becomes reliable at scale, while family captures task-dependent structure that varies by domain. Removing either degrades performance (Figure~\ref{fig:prior}, ablation insets), confirming that both global and task-specific structure are necessary for accurate model recommendation.

\subsection{Case Study: Cross-Domain Model Recommendation}
We further conduct two case studies on recently released benchmarks not in our training corpus, each probing a different aspect of \method: \textit{NGQA}~\citep{ngqa} tests whether \method produces useful task-specific recommendations that beat practical defaults, and \textit{RSVLM-QA}~\citep{rsvlmqa} tests whether its ranking is accurate beyond top-1 and generalizes to unseen candidates. The two benchmarks span text and vision-language modalities, providing a robust platform to examine cross-domain transfer.
% \textit{Case 1: NGQA (Text-based Reasoning).}

\begin{figure}[t]
    \centering
    \begin{subfigure}{0.48\textwidth}
        \centering
        \includegraphics[width=\linewidth]{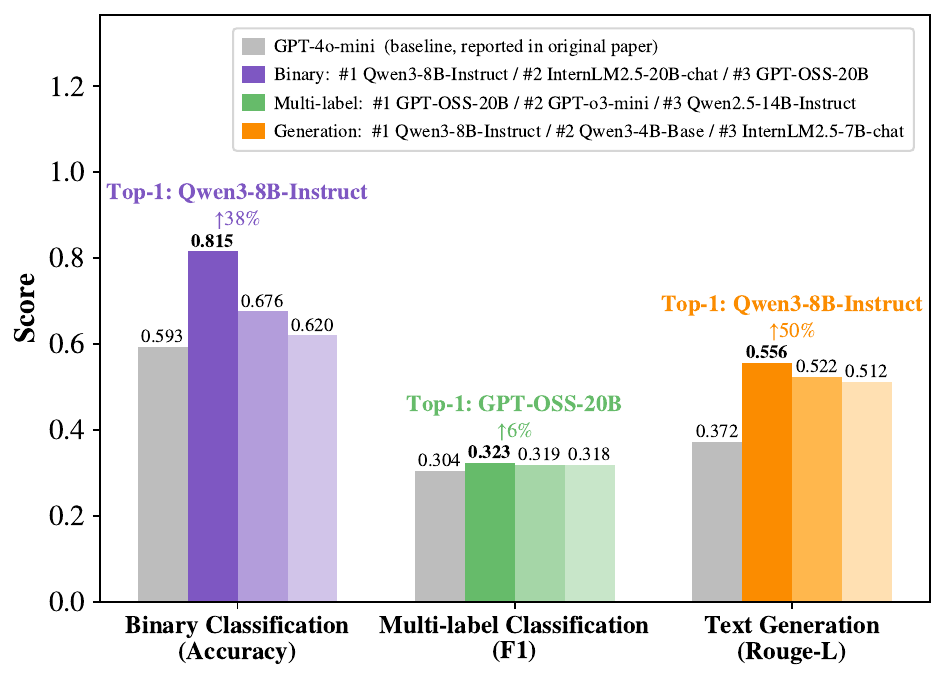}
    \end{subfigure}
    \hfill
    \begin{subfigure}{0.48\textwidth}
        \centering
        \includegraphics[width=\linewidth]{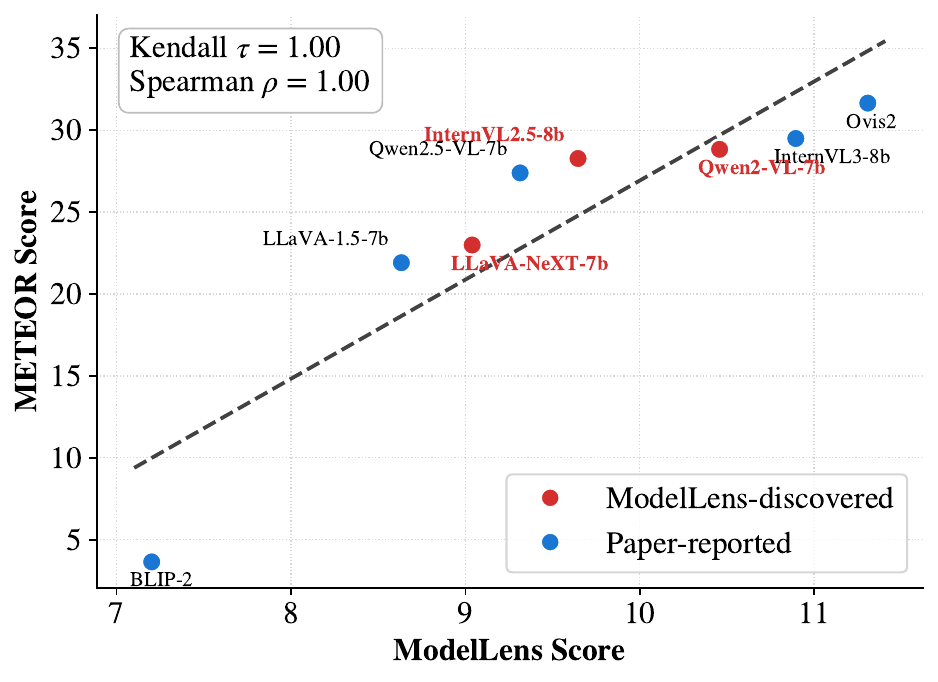}
    \end{subfigure}
    % \vspace{-4pt}
    \caption{Case studies on unseen datasets across domains. (Left) On NGQA, different tasks favor different models, and the recommendations consistently outperform the default baseline. (Right) On RSVLM-QA, predicted scores perfectly match empirical performance, recovering the full ranking and generalizing to additional competitive models not included in the original benchmark.}
    % \caption{Case studies on NGQA (left) and RSVLM-QA (right). \method outperforms the GPT-4o-mini default on NGQA and recovers the exact empirical ranking on RSVLM-QA.}
    \label{fig:case_study}
    % \vspace{-0.4cm}
\end{figure}

\textbf{Case 1: NGQA (Text-based Reasoning).}
NGQA spans three tasks over nutritional knowledge: binary classification, multi-label classification, and free-form text generation. We construct a controlled pool of models under 20B parameters (matching the implied scale of the default \texttt{GPT-4o-mini}). As shown in Figure~\ref{fig:case_study}(left), the optimal model varies across tasks, and the \method recommended top-1 consistently outperforms the default \texttt{GPT-4o-mini} in all settings, indicating that model suitability is task-dependent even within a single dataset, and \method captures this variation rather than committing to a single fixed choice.
% \textit{Case 2: RSVLM-QA (Vision-Language Understanding).}

\textbf{Case 2: RSVLM-QA (Vision-Language Understanding).}
We rank eight comparable-scale (7B–8B) vision-language models on the RSVLM-QA captioning subset, of which five appear in the original benchmark and three are surfaced by \method but not evaluated there. Figure~\ref{fig:case_study}(right) plots \method scores against METEOR. Our method recovers the exact empirical ranking ($\tau$=$\rho$=1.00), with \texttt{Ovis2} correctly identified as the best (METEOR = 31.65). Crucially, the three discovered models that are absent from the original benchmark fall precisely on the empirical regression trend, indicating that \method not only orders enumerated candidates correctly but also generalizes to identify additional competitive models without any direct evaluation.

%% file: sections/06conclusion.tex
\section{Conclusion}
We studied \emph{model recommendation in the wild}, the problem of identifying suitable models for a target task at the scale of today's open-source ecosystem. We introduced \method, a metric-aware ranking framework that learns directly from large-scale model–dataset–metric interactions and generalizes zero-shot to unseen models and datasets. On a benchmark of 1.62M evaluation records spanning 47K models and 9.6K datasets, \method surpasses both metadata-only and forward-pass-based transferability baselines without ever running a candidate on the target task, and its recommended Top-K pools translate into 21–81\% average gains across 5 representative routing methods. 
% Beyond ranking, \method yields a learned capability profile for each of the 47K models in our corpus, supporting analyses of strengths, blind spots, and family-level trends.
Beyond ranking, \method yields a learned capability profile for each of the 47K models in our corpus, supporting downstream analyses of model strengths, blind spots, and family-level trends. More broadly, our results suggest that the heterogeneous records accumulating in public leaderboards form a learnable capability atlas in their own right, a foundation layer for routing, ensembling, and model search as the open-model ecosystem expands.

%% file: sections/99appendix.tex
\section{Appendix}

\subsection{Appendix Overview}
\noindent

This appendix provides additional details, analyses, and reproducibility information for \method.
We organize the supplementary material as follows.

\begin{figure}
    \centering
    \includegraphics[width=\linewidth]{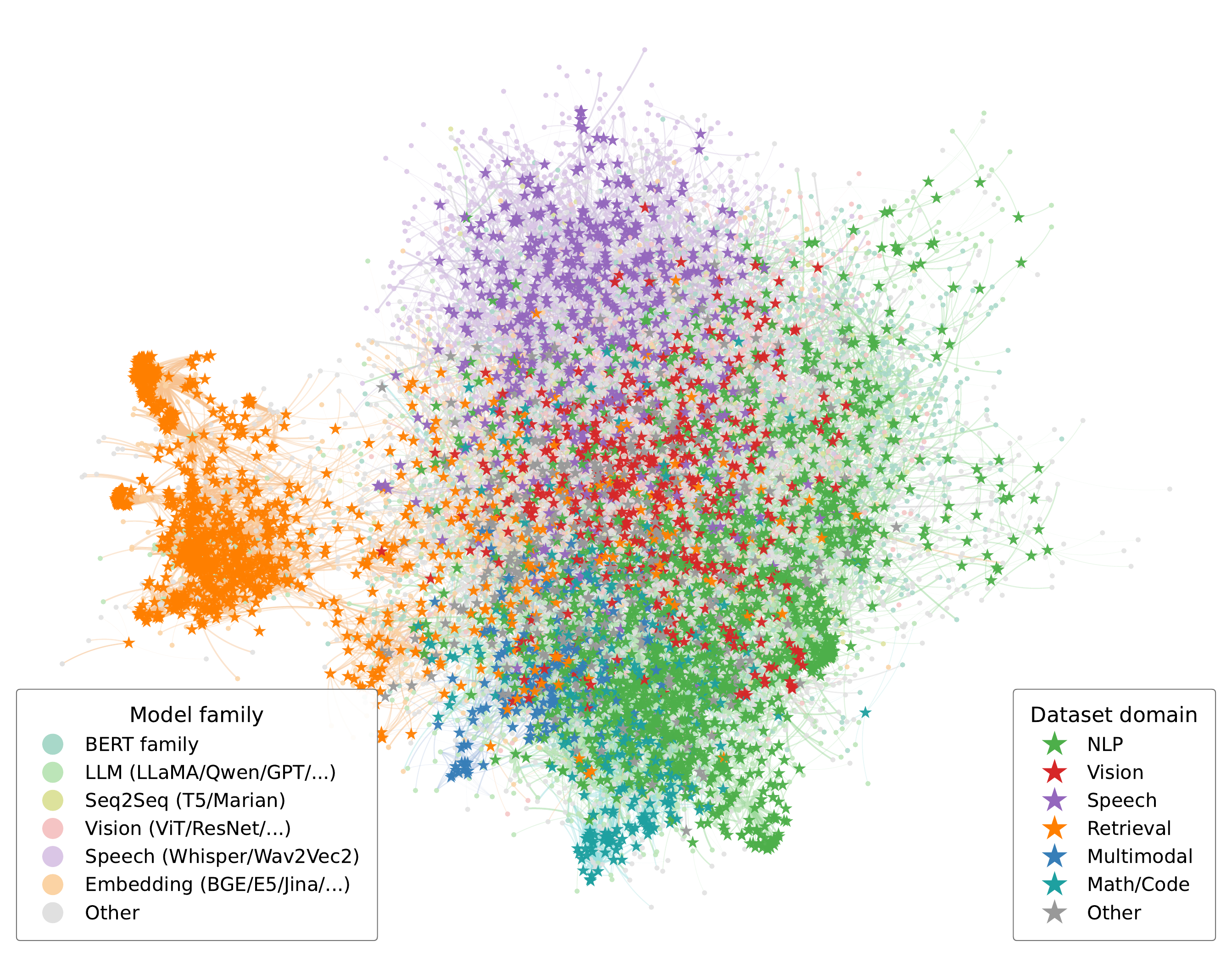}
    \caption{Visualization of the learned model--dataset embedding space trained on interaction data.Each point corresponds to either a model or a dataset, projected into a shared latent space learned from large-scale performance interactions. Compared to semantic-only representations, the learned space exhibits clear functional organization, where models and datasets cluster according to their task characteristics (e.g., NLP, vision, multimodal).This indicates that the model captures performance-aware relationships beyond surface-level similarity, enabling more meaningful grouping of models and datasets for downstream recommendation.}
    \label{fig:trained_interaction_atlas}
\end{figure}

\begin{figure}
    \centering
    \includegraphics[width=\linewidth]{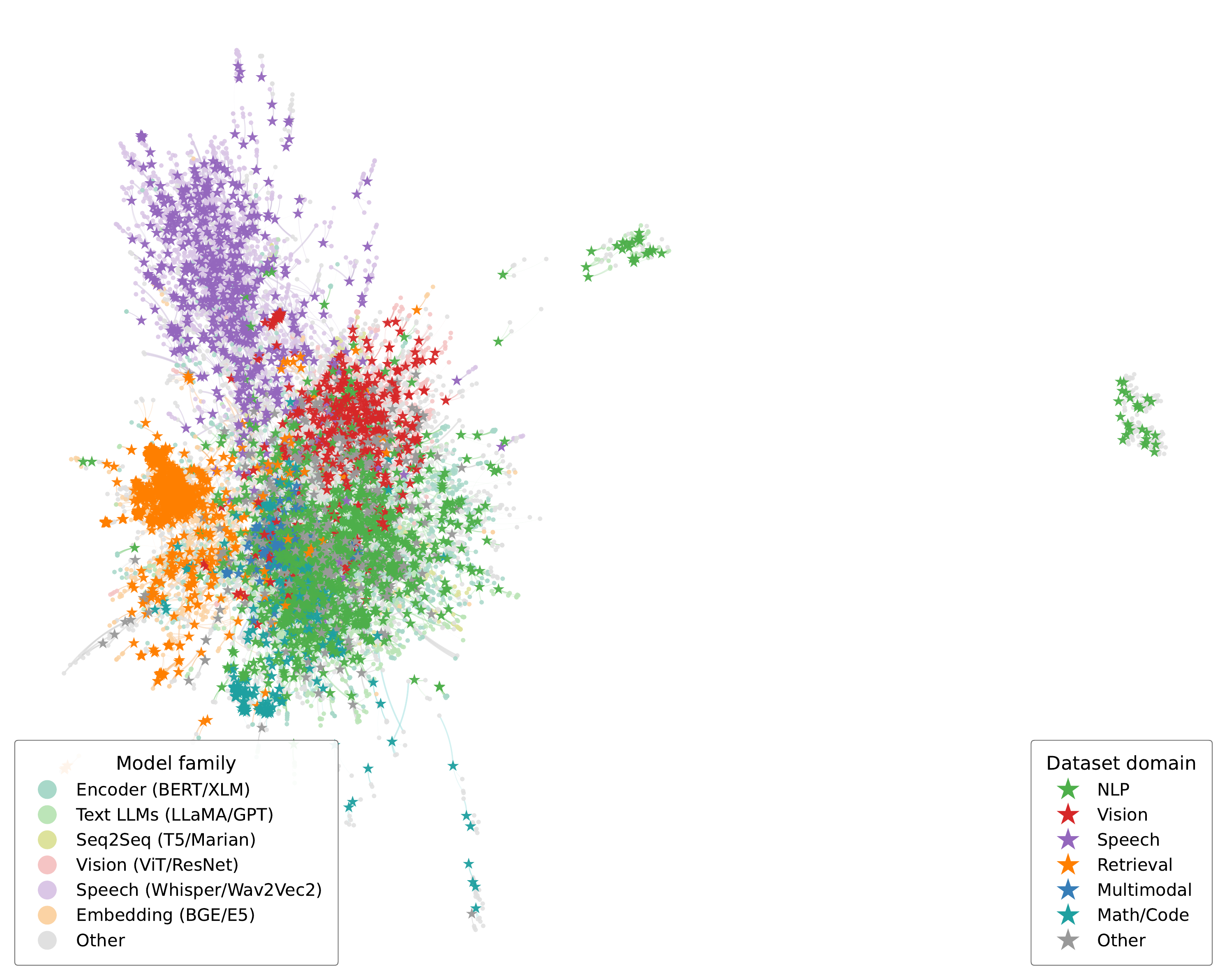}
    \caption{Visualization of the model--dataset embedding space constructed using semantic (content-based) features only. The embedding is derived from textual descriptions and metadata without leveraging performance interactions. Unlike the interaction-trained space, this representation shows limited structural organization, with different task domains intermingled and no clear clustering patterns. This highlights the limitation of relying solely on semantic similarity for model recommendation, as it fails to capture performance-relevant relationships between models and datasets.}
    \label{fig:semantic_only_atlas}
\end{figure}

\begin{appendixcard}{A. Learned Embedding Space}{~\Cref{fig:trained_interaction_atlas}--\Cref{fig:semantic_only_atlas}}
Visualizations of the interaction-trained and semantic-only model--dataset embedding spaces, highlighting how performance interactions induce more functional organization than textual similarity alone.
\end{appendixcard}

\begin{appendixcard}{B. Related Work and Comparison}{~\Cref{apx:related_detail}, ~\Cref{tab:matrix_completion_baselines}}
Detailed discussion of model profiling, transferability estimation, model routing, AutoML, and matrix-completion baselines, together with additional baseline results.
\end{appendixcard}

\begin{appendixcard}{C. Dataset Construction}{~\Cref{sec:dataset}, ~\Cref{tab:dataset_summary}}
Additional statistics and preprocessing details for the \emph{Model Recommendation in the Wild} benchmark, including data sources, the three-tier HuggingFace extraction pipeline, interaction normalization, metadata construction, and split design.
\end{appendixcard}

\begin{appendixcard}{D. Implementation and Evaluation Details}{~\Cref{sec:implementation}}
Architecture, embedding dimensions, optimization settings, ranking losses, batch construction, evaluation metrics, and compute resources used in our experiments.
\end{appendixcard}

\begin{appendixcard}{E. Baseline Details}{~\Cref{sec:baseline_details}}
Additional implementation details for feature-based transferability methods, feature-free methods, practitioner strawmen, and evaluation metrics.
\end{appendixcard}

\begin{appendixcard}{F. Recommended Model Pools for Routing}{~\Cref{apx:pools}, ~\Cref{tab:popqa_replace}--\Cref{tab:pool_controls}}
Recommended replacement pools for PopQA, HotpotQA, MuSiQue, and Bamboogle under comparable inference-scale constraints, together with the raw dataset descriptions used as semantic inputs.
\end{appendixcard}

\begin{appendixcard}{G. Ablations, Priors, and Case Studies}{~\Cref{apx:feature_ablation}, ~\Cref{apx:prior}}
Additional ablation results, feature-attribution analysis, standardized advantage computation, learned size/family priors, and full case-study rankings.
\end{appendixcard}

\begin{appendixcard}{H. Unseen-Family Generalization}{~\Cref{app:family-holdout}}
Evaluation under a strict family-level hold-out protocol, where entire modern LLM families are excluded from training and only appear at test time. This section analyzes unseen-family generalization and transferable model--dataset compatibility beyond family-specific memorization.
\end{appendixcard}

\begin{appendixcard}{I. Limitations, Broader Impacts, Assets, and Reproducibility}{~\Cref{apx:limitation}}
Discussion of limitations, broader impacts, data and asset licenses, and reproducibility information.
\end{appendixcard}

\newpage 

\newpage 

\clearpage 

\subsection{Detailed Related Works and Baseline Comparison}
\label{apx:related_detail}

\subsubsection{Model Profiling}

A growing body of work shifts the focus from individual models to 
entire model populations, treating models themselves as a data modality to support downstream applications such as model discovery and selection. We group this literature into three complementary directions.

\textbf{Weight space learning.}
The earliest direction studies models directly through their parameters. Unterthiner et al.~\citep{unterthiner2020predicting} showed that simple statistics of neural network weights suffice to predict model accuracy with high fidelity, even transferring across unseen datasets and architectures. Martin et al.~\citep{martin2021predicting} extended this to large pretrained models through a heavy-tailed self-regularization perspective, and Schürholt et al.~\citep{schurholt2022hyper} formalized 
the paradigm of \emph{weight space learning}, proposing self-supervised representations over model zoos that capture intrinsic properties such as accuracy and hyperparameters. More recent work scales this paradigm beyond curated zoos to large, heterogeneous repositories such as HuggingFace~\citep{wsl-huggingface}. While effective, all of these approaches require direct access to model weights and characterize models in isolation rather than their compatibility with specific tasks.

\textbf{Functional representations of models.}
A second line of work characterizes models through their functional behavior rather than their parameters. LLM DNA~\citep{llmdna} embeds models into a low-dimensional space based on their responses to probe inputs, enabling similarity analysis, clustering, and lineage inference. This approach avoids direct reliance on weight access and captures behavioral characteristics of models.
However, such methods typically require multiple forward passes per model, which limits scalability to large candidate pools. In addition, the resulting representations reflect global model similarity rather than task-specific suitability, making them insufficient for direct model selection.

\textbf{Model ecosystem analysis.}
A third direction studies the structural organization of model repositories. Model Atlas~\citep{modelatlas} proposes representing large model collections as a graph, where nodes correspond to models and edges capture transformations such as fine-tuning, quantization, or merging. This framework enables applications such as model forensics, lineage recovery, and meta-ML analysis over large-scale model ecosystems. Model Lakes~\cite{modellakes} frames model attribution, versioning, search, and benchmarking over large model collections as data-management tasks, and ModelTables~\cite{modeltables} builds a corpus of 90K tables from model cards, READMEs, and papers for structured table retrieval.
While these approaches provide a structured view of model populations and support model discovery at the infrastructure level, they do not address the core decision problem of selecting models for a given task, nor do they predict task-specific performance.

\textbf{Our position.}
Unlike prior work that represents models in isolation, we learn model--dataset \emph{interactions} directly from large-scale leaderboard data, predicting task-aware rankings without model weights or forward passes. This applies to both open and closed models and scales to ecosystems with tens of thousands of candidates.

\subsubsection{Transferability Estimation and Model Selection}
\label{apx:te_relateworks}
Transferability estimation (TE) aims to predict how well a pre-trained model will perform on a target task without the prohibitive cost of full fine-tuning. \textit{Training-free methods} typically estimate transferability by performing a single forward pass on the target dataset to extract feature--label statistics. Early works such as H-Score~\cite{h-score} and NCE~\cite{nce} utilize information-theoretic measures, while LEEP~\cite{leep} and NLEEP~\cite{nleep} extend these concepts using soft-label distributions. LogME~\cite{logme} formulates transferability as a marginal likelihood problem, and subsequent research has introduced variants like PACTran~\cite{pactran}, OTCE~\cite{otce}, LFC~\cite{lfc}, and GBC~\cite{gbc} to address specific transfer scenarios.

\textit{Learning-based methods} move beyond static metrics by leveraging interaction patterns. For instance, Model-Spider~\cite{modelspider} employs a cross-attention meta-ranker over feature representations, and Know2Vec~\cite{know2vec} maps per-class statistics into a shared embedding space. While effective, these approaches are fundamentally limited by their reliance on target-task inference~\cite{modelspider, know2vec}. As the ecosystem expands to tens of thousands of models, running forward passes for every candidate becomes computationally infeasible~\cite{huggingface, openllmleaderboard}. \method diverges from this paradigm by predicting rankings directly from leaderboard interactions and structural metadata, while retaining the flexibility to incorporate forward-pass features as optional augmentations when compute allows.

\subsubsection{Model Routing and Adaptive Inference}
\label{app:related_routing}
Existing routing methods typically assume the candidate pool is predefined and relatively small~\cite{routereval}. However, in a heterogeneous model space, the quality of this "upstream" pool significantly impacts downstream routing efficiency. \method serves as a foundational step for these systems by producing high-quality, task-specific candidate sets at the dataset level, which can then be seamlessly consumed by instance-level routers~\cite{router-r1, routerdc}.

\subsubsection{AutoML and Surrogate Modeling}
\label{apx:automl_relateworks}
The challenge of model recommendation in the wild is closely related to Zero-shot AutoML~\cite{automl}. Methods like ZAP~\cite{zap} and TabPFN~\cite{tabpfn} utilize neural surrogates to predict performance across diverse tasks. Similarly, Optimus~\cite{optimus} leverages large language models for optimization modeling. However, these methods often struggle with the scale and modality heterogeneity inherent in modern model hubs. By incorporating structural priors inspired by neural scaling laws~\cite{kaplan2020scaling} and architectural family biases, \method explicitly reasons about model capacity~\cite{wen-etal-2025-thinkguard} and multimodal robustness~\cite{ModalityInterference, DBLP:journals/corr/abs-2512-02306}, enabling more robust generalization than traditional surrogate models.

\begin{table*}[t]
\centering
\caption{Comparison with matrix-completion and inductive recommendation baselines under performance completion and new dataset evaluation settings. Best results are in bold, and second-best results are underlined.}
\label{tab:matrix_completion_baselines}
\small
\resizebox{\textwidth}{!}{%
\begin{tabular}{>{\raggedright\arraybackslash}p{3.4cm} l c|ccc|ccc|ccc}
\toprule
\textbf{Setting} & \textbf{Method} & \textbf{$\tau_{w}$}
& \textbf{NDCG@1} & \textbf{Hit@1} & \textbf{Recall@1}
& \textbf{NDCG@10} & \textbf{Hit@10} & \textbf{Recall@10}
& \textbf{NDCG@30} & \textbf{Hit@30} & \textbf{Recall@30} \\
\midrule

\multirow{7}{*}{\makecell[l]{\textbf{Performance Completion} \\ (2967 datasets)}}
& \textbf{ModelLens}
& \textbf{0.868}
& \textbf{0.954} & \textbf{0.153} & \textbf{0.153}
& \textbf{0.967} & \textbf{0.521} & \textbf{0.452}
& \textbf{0.974} & \underline{0.840} & \textbf{0.764} \\
& TwoTowerCosine
& 0.765
& 0.900 & 0.093 & 0.093
& 0.918 & 0.434 & 0.341
& 0.933 & 0.632 & 0.629 \\
& Standardized Embedder~\citep{StandardizedEmbedder}
& 0.167
& 0.661 & 0.036 & 0.036
& 0.717 & 0.138 & 0.096
& 0.755 & 0.237 & 0.258 \\
& MF~\citep{mf}
& \underline{0.843}
& \underline{0.935} & \underline{0.126} & \underline{0.126}
& \underline{0.946} & 0.489 & \underline{0.424}
& \underline{0.956} & 0.781 & \underline{0.720} \\
& IMC~\citep{imc}
& 0.739
& 0.894 & 0.064 & 0.064
& 0.910 & 0.462 & 0.340
& 0.921 & 0.691 & 0.571 \\
& GIMC~\citep{gimc}
& 0.503
& 0.776 & 0.078 & 0.078
& 0.823 & 0.319 & 0.214
& 0.864 & 0.652 & 0.498 \\
& IGMC~\citep{igmc}
& 0.762
& 0.907 & 0.089 & 0.089
& 0.927 & \underline{0.520} & 0.403
& 0.940 & \textbf{0.843} & 0.665 \\

\midrule

\multirow{7}{*}{\makecell[l]{\textbf{New Datasets} \\ (2764 datasets)}}
& \textbf{ModelLens}
& \textbf{0.817}
& \textbf{0.908} & \textbf{0.171} & \textbf{0.171}
& \textbf{0.929} & \textbf{0.471} & \textbf{0.410}
& \textbf{0.943} & \textbf{0.855} & \textbf{0.672} \\
& TwoTowerCosine
& 0.798
& \underline{0.900} & \underline{0.097} & \underline{0.097}
& \underline{0.918} & 0.418 & 0.335
& \underline{0.935} & 0.631 & \underline{0.638} \\
& Standardized Embedder~\citep{StandardizedEmbedder}
& 0.346
& 0.668 & 0.042 & 0.042
& 0.718 & 0.147 & 0.098
& 0.758 & 0.260 & 0.257 \\
& MF~\citep{mf}
& 0.726
& 0.879 & 0.047 & 0.047
& 0.904 & 0.324 & 0.268
& 0.923 & 0.586 & 0.609 \\
& IMC~\citep{imc}
& 0.792
& 0.900 & 0.058 & 0.058
& 0.916 & \underline{0.448} & \underline{0.345}
& 0.926 & \underline{0.712} & 0.580 \\
& GIMC~\citep{gimc}
& 0.576
& 0.766 & 0.085 & 0.085
& 0.821 & 0.291 & 0.210
& 0.861 & 0.637 & 0.490 \\
& IGMC~\citep{igmc}
& \underline{0.800}
& 0.866 & 0.072 & 0.072
& 0.889 & 0.334 & 0.277
& 0.912 & 0.564 & 0.579 \\

\bottomrule
\end{tabular}
}
\end{table*}

\subsubsection{Matrix Completion in Recommender Systems}
From a collaborative filtering perspective, model recommendation can be formulated as a sparse matrix completion problem, where the target is to populate a performance matrix $Y \in \mathbb{R}^{N \times T}$ representing $N$ models and $T$ datasets. Traditional Matrix Factorization (\texttt{MF})~\cite{mf} techniques excel at capturing latent interactions but are fundamentally transductive, relying on fixed identity (ID) embeddings that cannot generalize to the "cold-start" scenario of newly released models or datasets. 
To address this limitation, Inductive Matrix Completion (\texttt{IMC})~\cite{imc} and its Goal-directed \texttt{GIMC}~\cite{gimc} variants incorporate side information to enable prediction for unseen entities. Furthermore, \texttt{IGMC}~\cite{igmc} utilizes graph neural networks to learn inductive representations from local subgraphs, providing a powerful framework for reasoning over sparse interaction data. In the specific domain of model selection, the \texttt{TwoTowerCosine} architecture—often paired with \texttt{Task2Vec}~\cite{task2vec} for generating task-specific embeddings—has become a standard for aligning model capabilities with task requirements in a shared latent space. Additionally, the \texttt{Standardized Embedder}~\cite{StandardizedEmbedder} framework focuses on the fundamental scalability of these selections, offering asymptotically fast updates essential for maintaining myriads of models.

\method advances these inductive paradigms by introducing a dual-pathway scoring function that decomposes performance into a structural prior and a residual interaction term. While existing inductive methods like IMC rely heavily on side features, \method draws inspiration from \texttt{DropoutNet}~\cite{dropoutnet} and employs a unique \textit{ID-dropout} mechanism. This mechanism induces a \textit{dual-mode training regime}: a "memorization mode" that utilizes learned ID embeddings for high-fidelity ranking of seen models, and a "semantic mode" that forces the model to leverage name, description, and structural attributes for zero-shot generalization to unseen entities. Unlike traditional DropoutNet applications in generic recommendation, our framework specifically integrates this mechanism with a structural prior derived from neural scaling laws~\cite{kaplan2020scaling} and architectural families By grounding the shared representation in absolute performance magnitudes via an auxiliary pointwise loss, \method ensures a more robust calibration than traditional ranking-only matrix completion.

\clearpage

\begin{table*}[t]
\centering
\caption{Summary of the \emph{Model Recommendation in the Wild} dataset.}
\label{tab:dataset_summary}
\small
\resizebox{\textwidth}{!}{%
\begin{tabular}{lcccc}
\toprule
\textbf{Category} & \textbf{Attribute} & \textbf{Value} & \textbf{Description} & \textbf{Notes} \\
\midrule
\multirow{4}{*}{Scale}
& \# Models & 47,062 & Unique pretrained models & Across multiple domains \\
& \# Datasets & 9,682 & Distinct evaluation datasets & Includes vision, NLP, speech \\
& \# Tasks & 2,551 & Task categories & Unified taxonomy \\
& \# Metrics & 8420 & Unique evaluation metrics & Unified taxonomy \\
& \#Interactions & 1,623,284 & Model--dataset--metric evaluation pairs & After deduplication \\
\midrule
\multirow{3}{*}{Sources}
& HuggingFace & 1.64M (raw) & Model hub extraction & Gold/Silver/Bronze pipeline \\
& Open LLM Leaderboard & 147K & LLM benchmarks & Dense evaluation \\
& PapersWithCode & 10.8K & SOTA results & Sparse but diverse \\
\midrule
\multirow{3}{*}{Representation}
& Dataset Embedding & 1536-d & Text encoder (Text-Embedding-3-small) & Description-based \\
& Model Size & 21 buckets & Parameter discretization & Structural prior \\
& Model Family & 348 categories & Architecture grouping & e.g., LLaMA, ViT \\
\midrule
\multirow{3}{*}{Splits}
& Train & 1.27M & Training set & Stratified by dataset \\
& Validation & 168K & Hyperparameter tuning & -- \\
& Test & 187K & In-distribution evaluation & -- \\
\midrule
\multirow{2}{*}{OOD Setting}
& Held-out Datasets & 609 & Unseen datasets & No overlap with train \\
& OOD Interactions & 746K & Evaluation pairs & Generalization test \\
\bottomrule
\end{tabular}
}
\end{table*}

\subsection{Dataset Construction Details}
\label{sec:dataset}

We provide additional details of the data collection and preprocessing pipeline for the \emph{Model Recommendation in the Wild} benchmark. 
A summary of dataset statistics, sources, representations, and splits is provided in Table~\ref{tab:dataset_summary}.

\textbf{Data Sources}
We aggregate model--dataset performance interactions from three complementary sources:

\textit{HuggingFace Model Hub.}
We develop a three-tier extraction pipeline to collect evaluation results from model repositories:
(i) structured YAML results (\texttt{.eval\_results/}), 
(ii) standardized \texttt{model-index} metadata in model cards, and 
(iii) Markdown tables in README files, which are parsed into structured tuples using an LLM-based extractor.
This process yields 1.64M raw interactions across diverse pipeline tags.

\textit{Open LLM Leaderboard.}
We incorporate 147K evaluation interactions from 3,495 large language models across 43 benchmark datasets.

\textit{PapersWithCode.}
We include 10.8K interactions covering 5,443 models and 2,070 datasets from publicly reported results.

\textbf{Data Processing}
All interactions are unified into a standard tuple $(m, d, t, \mu, v)$, representing the performance of model $m$ on dataset $d$ under task $t$ and metric $\mu$.

We apply standard preprocessing steps including deduplication across sources, normalization of dataset and metric names, and filtering of incomplete or inconsistent entries. 
After processing, the dataset contains 1.62M interactions.

\textbf{Representation and Splits}
We encode dataset descriptions using pretrained text embeddings and incorporate model metadata such as parameter size and model family as structural features. 
Dataset-level splits are constructed via stratified sampling, and a cold-start setting is created by holding out a subset of datasets that do not appear during training. For unseen-model evaluation, held-out models are partitioned temporally according to their public release timestamps, simulating a realistic open-world setting in which newly released models must be recommended without prior interaction observations. For HuggingFace models, release timestamps are determined using the earliest public repository creation time or first available model-card timestamp.

\textbf{Discussion}
Compared to prior benchmarks, our dataset is distinguished by its scale, heterogeneity across domains and modalities, and its grounding in real-world, noisy reporting practices from model repositories.

\newpage
\subsection{Implementation Details}
\label{sec:implementation}
Our recommendation framework is implemented in PyTorch and trained with a joint listwise--pairwise ranking objective.
Unless otherwise specified, all experiments use the same backbone architecture, embedding configuration, and optimization settings across datasets and evaluation regimes.
We describe the main implementation details below.

\subsubsection{Model Architecture and Embeddings}

\textbf{Embedding configuration.}
The default full-feature configuration uses the dimensions in Table~\ref{tab:emb-dims}. Model and dataset description embeddings are
pre-computed using \texttt{text-embedding-3-small} (dim $1536$) and cached before training. These embeddings remain frozen during optimization. Hashed model-name tokens are represented with a trainable embedding table of dimension $512$. Discrete metadata features, including model size bucket, model family, task ID, and dataset ID, are represented with lightweight learnable embeddings. The final scoring head is implemented as a two-layer MLP with hidden width $512$ and dropout rate $0.02$. To improve generalization to unseen models and datasets, we additionally apply learned-ID dropout ($p=0.1$) on the model and dataset ID embeddings during training.

\begin{table}[h]
\centering
\small
\begin{tabular}{lc}
\toprule
Component & Dimension \\
\midrule
Model description embedding (frozen) & $1536$ \\
Model name token embedding (learned) & $512$ \\
Model size bucket embedding & $64$ \\
Model family embedding & $64$ \\
Dataset description embedding (frozen) & $1536$ \\
Dataset ID embedding (learned) & $256$ \\
Task ID embedding & $256$ \\
MLP hidden width & $512$ \\
\bottomrule
\end{tabular}
\caption{Embedding and hidden dimensions used in the full-feature ranker.}
\label{tab:emb-dims}
\end{table}

\subsubsection{Optimization and Training}

\textbf{Optimizer and regularization.}
All models are trained using \texttt{AdamW}
with learning rate $1\times10^{-3}$ and weight decay $1\times10^{-4}$.
We do not use a learning-rate scheduler; instead, training is controlled through early stopping with patience of $20$ epochs based on validation weighted Kendall's $\tau$. Gradients are clipped to a global $\ell_2$ norm of $5.0$ at every step.

\textbf{Batch construction.}
Listwise batches contain $B_{\text{list}}=8$ datasets, each expanded into its full candidate ranklist (typically $20$--$200$ models). Pairwise batches contain $B_{\text{pair}}=1024$ anchor--negative pairs. The listwise and pairwise loaders are interleaved during training so that each epoch terminates when the listwise loader is exhausted.

\textbf{Random seeds.} 
All experiments are run with three random seeds. Variation across seeds is small relative to the gaps between methods, so we report results from a single run for readability.

\subsubsection{Ranking Objectives}

\textbf{Target normalization.}
For each $(\text{task},\text{dataset})$ group, we sort candidate models by their raw evaluation metric and apply z-score normalization within the group. This normalization mitigates heterogeneity across metrics such as accuracy, F1, EM, and MMLU score, allowing the model to learn relative ranking signals instead of absolute metric values.

\textbf{Pair construction.}
For the pairwise objective, each anchor corresponds to a position
$i\in\{0,\ldots,M-2\}$ in the ground-truth ranking.
One negative is sampled uniformly from lower-ranked positions
$\{i+1,\ldots,M-1\}$.

\textbf{Joint ranking objective.}
The final ranker is trained with a joint objective:
\[
\mathcal{L}
=
\lambda_{\text{list}}\mathcal{L}_{\text{list}}
+
\lambda_{\text{pair}}\mathcal{L}_{\text{pair}}
+
\lambda_{\text{point}}\mathcal{L}_{\text{point}},
\]

where
$\lambda_{\text{list}}=0.5$,
$\lambda_{\text{pair}}=1.0$,
and
$\lambda_{\text{point}}=0.1$.

\textbf{Listwise objective.}
For a ranklist of length $M$ with predicted scores
$s_1,\ldots,s_M$ sorted by ground-truth rank, the listwise loss is:
\[
\mathcal{L}_{\text{list}}
=
\frac{1}{M}
\sum_{i=1}^{M}
\left[
\log
\sum_{j=i}^{M}\exp(s_j/\tau)
-
s_i/\tau
\right],
\]

where the temperature is set to $\tau=10$.

\textbf{Pairwise objective.}
Given anchor and negative scores $(s_+,s_-)$, we use the standard
Bayesian Personalized Ranking (BPR) objective:
\[
\mathcal{L}_{\text{pair}}
=
\mathbb{E}
\left[
-\log\sigma(s_+-s_-)
\right].
\]

\textbf{Pointwise auxiliary objective.}
Both branches additionally regress the predicted z-score against the
normalized ground-truth value using an MSE loss. This auxiliary objective stabilizes optimization during early training.

\subsubsection{Evaluation Protocol}
\textbf{Ranking metrics.}
For each $(\text{task},\text{dataset})$ group, we score all candidate models and compare the predicted ranking against the ground-truth ranking induced by held-out leaderboard metrics.
Our primary metric is weighted Kendall's $\tau$, averaged across groups with weight $1/M$ to avoid domination by large candidate pools.
We additionally report NDCG@$k$, Hit@$k$, and Recall@$k$
for $k\in\{1,10,30,50\}$.
Groups with fewer than $k$ candidates are excluded from the corresponding @$k$ metric.

\textbf{Model selection.}
The checkpoint reported in the main paper is selected based on the best validation weighted Kendall's $\tau$. Checkpoints are evaluated and saved automatically at every epoch whenever validation performance improves.

\subsubsection{Compute Resources}
\textbf{Hardware.}
The main model is trained using 1 A6000 GPU with PyTorch DistributedDataParallel. A full training run typically converges within approximately 6--8 hours wall-clock time due to early stopping, while evaluation on the complete test grid requires less than 5 minutes.

\textbf{Scalability.}
Our framework operates entirely on leaderboard interactions and metadata,without requiring downstream model execution or fine-tuning during recommendation inference. Recommendation complexity scales linearly with the number of candidate models for a given dataset.

% * optimizer
% * scheduler
% * negative sampling
% * pair construction
% * listwise/pairwise implementation
% * embedding dimension
% * batch size
% * GPUs

\newpage
\subsection{Baseline Details}\label{sec:baseline_details}

\textbf{Feature-based transferability methods.}
These methods compute a scalar score for each candidate model given a target dataset, based on feature or label statistics extracted from a forward pass. We consider both training-free and learning-based approaches. Training-free methods compute a scalar score per model from feature or label statistics obtained via a single forward pass: \texttt{H-Score}~\citep{h-score}, \texttt{NCE}~\citep{nce}, \texttt{LEEP}~\citep{leep}, \texttt{NLEEP}~\citep{nleep}, \texttt{LogME}~\citep{logme}, \texttt{PACTran}~\citep{pactran}, \texttt{OTCE}~\citep{otce}, \texttt{LFC}~\citep{lfc}, and \texttt{GBC}~\citep{gbc}. Learning-based methods improve ranking quality by modeling interactions between model features and target data: \texttt{Model-Spider}~\citep{modelspider} is a cross-attention meta-learner over heterogeneous feature representations extracted on the target dataset, and \texttt{Know2Vec}~\citep{know2vec} maps per-class feature statistics and task queries into a shared embedding space. We follow the standard evaluation protocol of Zhang et al.~\citep{modelspider}.

\textbf{Feature-free methods.}
These methods do not require running models on the target dataset, and instead rely on dataset metadata or learned model–dataset interactions. \texttt{Task2Vec}~\citep{task2vec} embeds datasets via the Fisher Information of a probe network and transfers rankings from the nearest training datasets. \texttt{ZAP}~\citep{zap} is a neural surrogate predicting model performance from model and dataset features. Both were originally designed for small curated pools; we use the original implementations of both methods, restricting their inputs to entities present in our benchmark.

\textbf{Practitioner strawmen.}
When no recommendation tool is available, practitioners commonly fall back on simple heuristics. We include two such baselines: \texttt{Model Size} ranks candidates purely by parameter count (reflecting the heuristic that larger models perform better), and \texttt{Model Popularity} ranks them by recent HuggingFace download counts (reflecting community-aggregated quality signals).

\textbf{Evaluation metrics.}
We use Kendall's weighted \textit{$\tau_w$}~\citep{kendalltau} rather than standard $\tau$ because misorderings near the top of the ranking matter more than those at the tail in practice — a recommender that confuses ranks 1 and 2 is far more harmful than one confusing ranks 100 and 101. We complement it with three top-$K$ metrics that capture different aspects of recommendation quality: \textit{Hit@$K$} measures whether any truly top-ranked model appears in the top-$K$, \textit{NDCG@$K$} measures position-weighted relevance, and \textit{Recall@$K$} measures coverage of competitive models. All metrics are computed per dataset and averaged across the test set.

\newpage

\subsection{Recommended Model Pools for Routing}
\label{apx:pools}

We present the model pools recommended by \method for several representative question answering benchmarks, including PopQA, HotpotQA, MuSiQue, and Bamboogle, in Tables~\ref{tab:popqa_replace}--\ref{tab:bamboogle_replace}. 
For each dataset, \method generates a replacement pool under comparable inference scale constraints, where candidate models are selected based on predicted compatibility with the target dataset semantics rather than direct evaluation on the benchmark itself. 
The resulting pools illustrate how \method adapts model selection to different reasoning demands, factual recall requirements, and robustness characteristics across datasets.

To improve reproducibility, we additionally provide in Table~\ref{tab:dataset_descriptions} the exact raw dataset descriptions used as semantic metadata inputs for pool generation. 
These descriptions are directly consumed by the routing framework to construct dataset representations and infer capability requirements for model recommendation, without manual feature engineering or benchmark-specific heuristics.

\begin{table*}[t]
\centering

\begin{minipage}{0.95\textwidth}
\centering
\captionof{table}{Model pool replacement for PopQA~\citep{popqa} under comparable inference scale.}
\label{tab:popqa_replace}
\small
\setlength{\tabcolsep}{6pt}
\renewcommand{\arraystretch}{0.95}
\begin{tabular}{l c c c l}
\toprule
\textbf{Original} & \textbf{Scale} & $\rightarrow$ & \textbf{Scale} & \textbf{Selected} \\
\midrule
LLaMA-3.1-70B & $\approx 70$B & $\rightarrow$ & $\approx 70$B & LLaMA-3.3-70B \\
Mixtral-8x22B & $\approx 39$B & $\rightarrow$ & $\approx 20$B & GPT-OSS-20B \\
Gemma-2-27B & $\approx 27$B & $\rightarrow$ & $\approx 14$B & Mixtral-8x7b-v0.1 \\
LLaMA-3.1-8B & $\approx 8$B & $\rightarrow$ & $\approx 8$B & LLaMA-3-8B \\
Qwen2.5-7B & $\approx 7$B & $\rightarrow$ & $\approx 4$B & Gemma-3n-E4B \\
Mistral-7B & $\approx 7$B & $\rightarrow$ & $\approx 3$B & Trinity-Mini-Base \\
\bottomrule
\end{tabular}
\end{minipage}

\vspace{0.35cm}

\begin{minipage}{0.95\textwidth}
\centering
\captionof{table}{Model pool replacement for HotpotQA~\citep{hotpotqa} under comparable inference scale.}
\label{tab:hotpotqa_replace}
\small
\setlength{\tabcolsep}{6pt}
\renewcommand{\arraystretch}{0.95}
\begin{tabular}{l c c c l}
\toprule
\textbf{Original} & \textbf{Scale} & $\rightarrow$ & \textbf{Scale} & \textbf{Selected} \\
\midrule
LLaMA-3.1-70B & $\approx 70$B & $\rightarrow$ & $\approx 70$B & LLaMA-3.3-70B \\
Mixtral-8x22B & $\approx 39$B & $\rightarrow$ & $\approx 20$B & GPT-OSS-20B \\
Gemma-2-27B & $\approx 27$B & $\rightarrow$ & $\approx 17$B & Llama-4-Maverick \\
LLaMA-3.1-8B & $\approx 8$B & $\rightarrow$ & $\approx 8$B & Qwen3-8b \\
Qwen2.5-7B & $\approx 7$B & $\rightarrow$ & $\approx 7$B & LLaMA-3.1-8B \\
Mistral-7B & $\approx 7$B & $\rightarrow$ & $\approx 3$B & Trinity-Mini-Base \\
\bottomrule
\end{tabular}
\end{minipage}

\vspace{0.35cm}

\begin{minipage}{0.95\textwidth}
\centering
\captionof{table}{Model pool replacement for Musique~\citep{musique} under comparable inference scale.}
\label{tab:musique_replace}
\small
\setlength{\tabcolsep}{6pt}
\renewcommand{\arraystretch}{0.95}
\begin{tabular}{l c c c l}
\toprule
\textbf{Original} & \textbf{Scale} & $\rightarrow$ & \textbf{Scale} & \textbf{Selected} \\
\midrule
LLaMA-3.1-70B & $\approx 70$B & $\rightarrow$ & $\approx 72$B & Qwen2.5-72B \\
Mixtral-8x22B & $\approx 39$B & $\rightarrow$ & $\approx 32$B & Kimi-K2.5 \\
Gemma-2-27B & $\approx 27$B & $\rightarrow$ & $\approx 17$B & Llama-4-Maverick \\
LLaMA-3.1-8B & $\approx 8$B & $\rightarrow$ & $\approx 8$B & Nemotron-H-8B-R \\
Qwen2.5-7B & $\approx 7$B & $\rightarrow$ & $\approx 7$B & Qwen2.5-7B \\
Mistral-7B & $\approx 7$B & $\rightarrow$ & $\approx 4$B & Gemma-3n-E4B \\
\bottomrule
\end{tabular}
\end{minipage}

\vspace{0.35cm}

\begin{minipage}{0.95\textwidth}
\centering
\captionof{table}{Model pool replacement for Bamboogle~\citep{bamboogle} under comparable inference scale.}
\label{tab:bamboogle_replace}
\small
\setlength{\tabcolsep}{6pt}
\renewcommand{\arraystretch}{0.95}
\begin{tabular}{l c c c l}
\toprule
\textbf{Original} & \textbf{Scale} & $\rightarrow$ & \textbf{Scale} & \textbf{Selected} \\
\midrule
LLaMA-3.1-70B & $\approx 70$B & $\rightarrow$ & $\approx 72$B & Qwen2.5-72B \\
Mixtral-8x22B & $\approx 39$B & $\rightarrow$ & $\approx 32$B & Kimi-K2.5 \\
Gemma-2-27B & $\approx 27$B & $\rightarrow$ & $\approx 22$B & Qwen3-235B-A22B \\
LLaMA-3.1-8B & $\approx 8$B & $\rightarrow$ & $\approx 8$B & Llama-3-8B \\
Qwen2.5-7B & $\approx 7$B & $\rightarrow$ & $\approx 7$B & Mistral-7B \\
Mistral-7B & $\approx 7$B & $\rightarrow$ & $\approx 3$B & Trinity-Mini-Base \\
\bottomrule
\end{tabular}
\end{minipage}

\end{table*}

\newpage

\begin{longtable}{p{1.8cm} p{2cm} p{11cm}}
\caption{
Raw dataset descriptions used as semantic inputs for model recommendation.
}
\label{tab:dataset_descriptions}
\\

\toprule
\textbf{Dataset} & \textbf{Task} & \textbf{Raw Description} \\
\midrule
\endfirsthead

HotpotQA
& Question Answering
& HotpotQA is a large-scale question answering dataset specifically designed to evaluate multi-hop reasoning and explainable question answering systems. Unlike single-document QA benchmarks, HotpotQA requires models to jointly reason over multiple documents to derive the correct answer, reflecting more complex and realistic information-seeking scenarios. The dataset is constructed from Wikipedia articles and contains over 110,000 question-answer pairs, each associated with multiple supporting documents. Crucially, HotpotQA provides explicit supporting fact annotations, identifying the exact sentences across different documents that are necessary for answering each question. This design enables fine-grained supervision not only on answer correctness but also on the reasoning process itself. \\

NQ
& Question Answering
& Natural Questions (NQ) is a large-scale open-domain question answering benchmark built from real user information-seeking queries issued to web search systems. Unlike synthetic or heavily curated QA datasets, NQ reflects authentic, noisy, and diverse question intents, making it well-suited for evaluating practical question answering performance in realistic settings. The dataset is grounded in Wikipedia documents and includes annotations for short and long answers, requiring models to identify both relevant evidence and precise answer spans. NQ emphasizes factual retrieval, answer extraction, and robustness to ambiguous or underspecified queries, and is widely used to assess whether language models can provide accurate and concise responses under real-world search-style distributions.
\\

PopQA
& Question Answering
& PopQA is an open-domain question answering dataset designed to stress-test parametric factual knowledge under popularity-aware distributions. It contains subject-relation-object style factual questions derived from knowledge graph triples, with accompanying metadata such as entity popularity that helps characterize long-tail difficulty. PopQA is particularly useful for evaluating whether models can answer less frequent or less memorized facts, rather than only high-frequency popular entities. For model routing, this benchmark provides a practical way to compare robustness across the popularity spectrum and to determine whether cheaper models suffice for common facts while stronger models are needed for harder, low-frequency knowledge queries.
\\

Musique
& Question Answering
& MuSiQue is a multi-hop question answering benchmark that explicitly targets compositional reasoning across multiple supporting paragraphs. Questions are built to require combining evidence from several facts, reducing shortcut opportunities that simpler single-hop datasets may allow. The benchmark includes decomposed reasoning supervision and supporting context structure, enabling analysis of both final-answer correctness and intermediate reasoning demands. In routing scenarios, MuSiQue is important because it distinguishes models that can perform deeper evidence composition and cross-document inference, helping select when high-capability models are necessary for complex reasoning-heavy QA requests.
\\

Bamboogle
& Question Answering
& Bamboogle is a challenge-style open-domain question answering dataset curated to include difficult, often deceptive or compositionally tricky questions that can expose weaknesses in shallow retrieval-and-match behavior. Many items require careful factual disambiguation, multi-step inference, or resistance to plausible but incorrect distractor knowledge. Compared with standard factual QA sets, Bamboogle places higher emphasis on robustness under adversarially challenging query formulations. For routing systems, Bamboogle is a useful stress benchmark to evaluate whether the router can identify hard queries and assign them to models with stronger reasoning and calibration, instead of over-allocating to lightweight models.
\\
\bottomrule
\end{longtable}

\clearpage
\subsection{Controls for Routing Pool Construction}
\label{app:pool_controls}

\begin{table}[h]
\centering
\caption{Alternative pool constructors on NQ under the Router-R1 protocol of Table~\ref{tab:routing}, with the same candidate universe and pool size. \textbf{Recency}: each original slot is replaced by the most recent same-family release in the same scale bucket (HuggingFace release dates, dense models matched on parameter count, MoE models on active parameters); ``kept'' marks slots with no eligible newer release. \textbf{ZAP}: pool selected by the strongest feature-free baseline of \Cref{tab:generalization}, listed by scale. \textbf{\method}: the recommended pool of \Cref{tab:model_replace}.}
\label{tab:pool_controls}
\small
\setlength{\tabcolsep}{4pt}
\begin{tabular}{llll}
\toprule
\textbf{Original (Router-R1)} & \textbf{Recency} & \textbf{ZAP} & \textbf{\method} \\
\midrule
LLaMA-3.1-70B & Llama-3.3-70B          & Llama-3.3-70B-Instruct       & LLaMA-3.3-70B \\
Mixtral-8x22B & Mixtral-8x22B (kept)   & Qwen2.5-32B-Instruct         & GPT-OSS-20B \\
Gemma-2-27B   & Gemma-3-27B-IT         & GPT-OSS-20B                  & Llama-4-Maverick \\
LLaMA-3.1-8B  & LLaMA-3.1-8B (kept)    & Mixtral-8x7B-Instruct-v0.1   & Nemotron-H-8B-R \\
Qwen2.5-7B    & Qwen3.5-9B             & Meta-Llama-3-8B-Instruct     & Qwen2.5-7B \\
Mistral-7B    & Ministral-8B-2410      & Qwen2.5-7B-Instruct          & Gemma-3n-E4B \\
\midrule
\textbf{NQ EM}            & 0.508 & 0.464 & \textbf{0.524} \\
\textbf{Active params}    & $\approx$161B & $\approx$150B & $\approx$\textbf{126}B \\
\midrule
\multicolumn{4}{l}{Original pool: NQ EM 0.388, $\approx$158B active parameters.} \\
\bottomrule
\end{tabular}
\end{table}

\subsection{Per-Task-Family Breakdown on New Datasets}
\label{app:task_family}

\Cref{tab:task_family} stratifies the 609 held-out datasets of~\Cref{tab:generalization} (New Datasets) by task family. \method improves over ZAP in every family, but the margin and the absolute ranking quality vary. It is strongest where candidate rankings follow transferable family- or scale-level regularities (speech recognition, bitext mining, classification), and weakest on embedding- and similarity-oriented tasks (reranking, retrieval, STS, pair classification), where rankings depend on narrow dataset-specific specialization that size and family priors do not capture; the same holds below $\sim$1B parameters, where the size prior is least reliable (Figure~\ref{fig:prior}).

\begin{table}[h]
\centering
\caption{Weighted Kendall $\tau_w$ per task family on the 609 held-out datasets, averaged over the datasets of each family; families with fewer than eight datasets are omitted. Bottom: fraction of held-out datasets on which the single best-performing model is ranked first, or within the top 10.}
\label{tab:task_family}
\small
\setlength{\tabcolsep}{6pt}
\begin{tabular}{lcccc}
\toprule
\textbf{Task family} & \textbf{\#Datasets} & \textbf{\method} & \textbf{ZAP} & $\boldsymbol{\Delta}$ \\
\midrule
Speech recognition    & 25  & 0.832 & 0.053  & +0.78 \\
Bitext mining         & 116 & 0.781 & 0.150  & +0.63 \\
Classification        & 148 & 0.779 & 0.407  & +0.37 \\
Question answering    & 13  & 0.677 & 0.075  & +0.60 \\
Clustering            & 34  & 0.673 & 0.368  & +0.31 \\
Pair classification   & 13  & 0.625 & 0.342  & +0.28 \\
STS                   & 56  & 0.615 & 0.375  & +0.24 \\
Text generation       & 84  & 0.602 & 0.099  & +0.50 \\
Retrieval             & 71  & 0.578 & 0.337  & +0.24 \\
Token classification  & 9   & 0.560 & $-$0.331 & +0.89 \\
Reranking             & 14  & 0.499 & 0.251  & +0.25 \\
\midrule
Best model at rank 1 (\%)  & 609 & \textbf{31.6} & 0.9  & +30.7 \\
Best model in top 10 (\%)  & 609 & \textbf{59.7} & 35.3 & +24.4 \\
\bottomrule
\end{tabular}
\end{table}

\newpage

\begin{table*}[t]
\caption{Model selection performance measured by MRR (for reference).}
\label{tab:mrr}
\centering
\small
\setlength{\tabcolsep}{4pt}
\resizebox{\textwidth}{!}{%
\begin{tabular}{l|ccccccccc}
\toprule
Method
& Aircraft & Cars & DTD & Pets & Flowers102 & Food101 & Country211 & EuroSAT & Avg. \\
\midrule

\multicolumn{10}{c}{\textbf{\textit{Feature-based Transferability Methods}}} \\
\midrule
H-Score      & 0.200 & 0.200 & 0.200 & 1.000 & 0.111 & 0.143 & 0.500 & 0.167 & 0.315 \\
NCE          & 0.333 & 0.250 & 1.000 & 0.333 & 0.333 & 0.111 & 0.500 & 0.143 & 0.375 \\
LEEP         & 0.250 & 1.000 & 0.500 & 0.200 & 0.500 & 0.167 & 0.333 & 0.111 & 0.383 \\
NLEEP        & 0.100 & 0.111 & 0.100 & 0.333 & 0.333 & 0.100 & 0.500 & 0.111 & 0.211 \\
LogME        & 0.125 & 0.111 & 0.100 & 0.500 & 0.125 & 0.200 & 0.500 & 0.167 & 0.229 \\
PACTran      & 0.125 & 0.100 & 0.100 & 1.000 & 0.167 & 0.125 & 0.125 & 1.000 & 0.343 \\
OTCE         & 0.200 & 0.100 & 0.100 & 1.000 & 0.143 & 0.111 & 0.125 & 1.000 & 0.347 \\
LFC          & 0.143 & 0.200 & 0.100 & 0.250 & 0.333 & 0.333 & 0.200 & 0.500 & 0.257 \\
GBC          & 0.100 & 0.100 & 0.333 & 0.200 & 0.167 & 0.333 & 0.333 & 0.111 & 0.210 \\
Model-Spider & 0.500 & 1.000 & 0.500 & 0.500 & 0.125 & 0.250 & 0.333 & 0.200 & 0.426 \\
Know2Vec     & 0.250 & 0.250 & 0.333 & 1.000 & 0.167 & 0.111 & 0.500 & 0.111 & 0.340 \\

\midrule
\multicolumn{10}{c}{\textbf{\textit{Feature-free Methods}}} \\
\midrule
Task2Vec & 0.143 & 0.111 & 0.200 & 0.125 & 0.333 & 0.250 & 0.500 & 0.250 & 0.239 \\
ZAP           & 0.100 & 0.200 & 1.000 & 0.111 & 0.100 & 1.000 & 0.250 & 0.200 & 0.370 \\

\midrule
\multicolumn{10}{c}{\textbf{\textit{Ours}}} \\
\midrule
Ours (Feature Free) & 0.250 & 0.333 & 1.000 & 0.250 & 0.250 & 0.500 & 1.000 & 0.333 & 0.490 \\
\textbf{Ours (Feature Aug.)}
& \textbf{0.500} & \textbf{1.000} & 0.333 & \textbf{1.000}
& \textbf{0.333} & \textbf{1.000} & \textbf{0.500} & \textbf{0.500}
& \textbf{0.635} \\

\bottomrule
\end{tabular}%
}
\end{table*}

\clearpage
\begin{table*}[t]
\caption{Ablation study on different loss combinations. Best results are in bold.}
\label{tab:ablation_loss}
\centering
\small
\resizebox{\textwidth}{!}{%
\begin{tabular}{l l c ccc ccc ccc}
\toprule
\textbf{Method} & \textbf{Loss} & \textbf{$\tau_{w}$} 
& \textbf{NDCG@1} & \textbf{Hit@1} & \textbf{Recall@1}
& \textbf{NDCG@10} & \textbf{Hit@10} & \textbf{Recall@10}
& \textbf{NDCG@30} & \textbf{Hit@30} & \textbf{Recall@30} \\
\midrule

Full (ensemble) & L+P+Pt 
& \textbf{0.745} 
& \textbf{0.910} & \textbf{0.266} & \textbf{0.266} 
& \textbf{0.951} & \textbf{0.456} & \textbf{0.303} 
& \textbf{0.962} & \textbf{0.666} & \textbf{0.631} \\

NoPointWise & L+P 
& \underline{0.728} 
& \underline{0.897} & \underline{0.126} & \underline{0.126} 
& \underline{0.935} & \underline{0.419} & \underline{0.300} 
& \underline{0.950} & 0.593 & \underline{0.620} \\

NoPairWise & L+Pt 
& 0.703 
& 0.896 & 0.080 & 0.080
& 0.930 & 0.405 & 0.284 
& 0.942 & \underline{0.631} & 0.582 \\

NoListWise & P+Pt 
& 0.632 
& 0.892 & 0.223 & 0.223
& 0.906 & 0.322 & 0.241 
& 0.912 & 0.475 & 0.410 \\

OnlyPairWise & P 
& 0.591 
& 0.885 & 0.015 & 0.015 
& 0.900 & 0.308 & 0.208 
& 0.908 & 0.473 & 0.431 \\

OnlyListWise & L 
& 0.649 
& 0.891 & 0.013 & 0.013
& 0.901 & 0.294 & 0.223 
& 0.924 & 0.456 & 0.442 \\

\bottomrule
\end{tabular}%
}
\end{table*}

\begin{figure}[t]
    \centering
    \begin{subfigure}{0.48\textwidth}
        \centering
        \includegraphics[width=\linewidth]{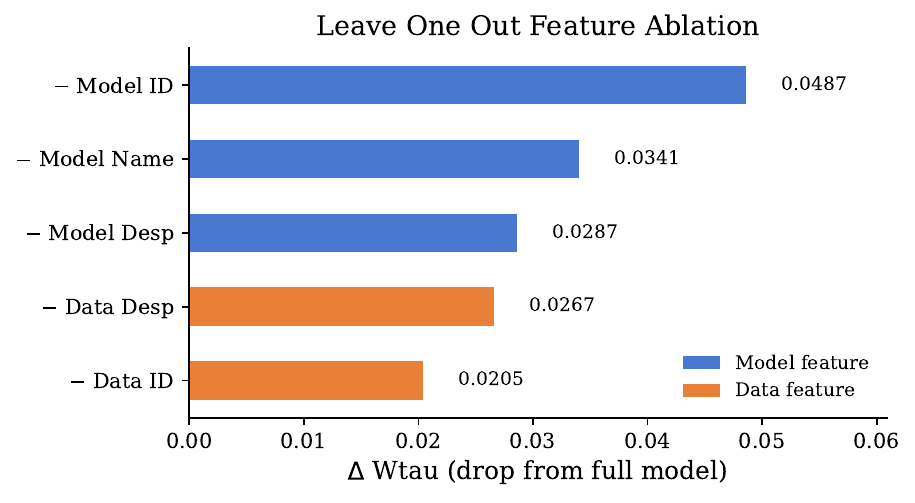}
    \end{subfigure}
    \hfill
    \begin{subfigure}{0.48\textwidth}
        \centering
        \includegraphics[width=\linewidth]{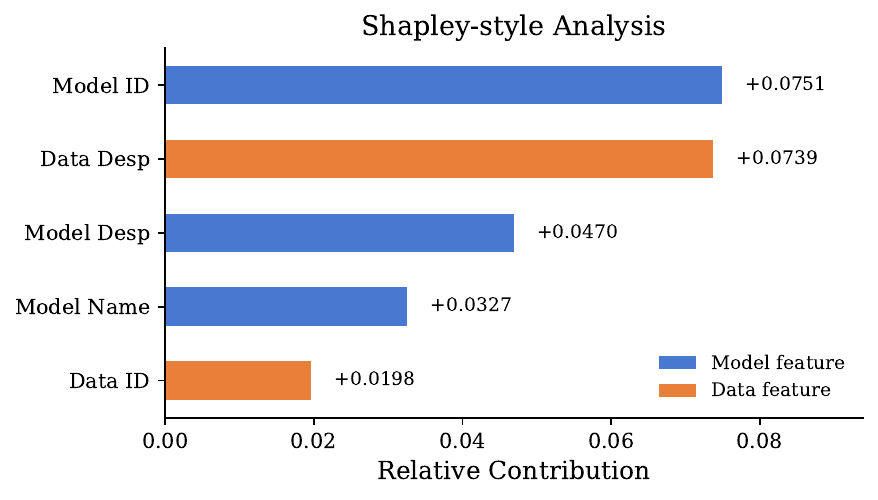}
    \end{subfigure}
    \caption{Comparison of ablation results and feature importance analysis.}
    \label{fig:feature_ablation}
\end{figure}

\subsection{Feature Ablation and Results}
\label{apx:feature_ablation}
We investigate the contribution of model-side (model ID, name, description) and dataset-side (data ID, description) features using two complementary attribution views, as illustrated in Figure~\ref{fig:feature_ablation}.
LOO drops quantify the performance degradation when each feature is removed from the full model, capturing its contribution under a fixed context. In contrast, Shapley-style analysis~\citep{shapley} measures the average marginal contribution of each feature across feasible feature subsets, providing a context-agnostic attribution.
Comparing LOO and Shapley estimates reveals strong feature interactions. 
In particular, Model Name and Model Desp exhibit pronounced redundancy: each contributes significantly in isolation, but their marginal gains shrink when combined, indicating overlapping model-identifying semantics that are obscured under LOO but captured by Shapley attribution.
Overall, while LOO highlights feature importance in the presence of all signals, Shapley-style analysis uncovers complementary contributions and redundancy across features.

\newpage
\subsection{Unseen-Family Generalization}
\label{app:family-holdout}
A practical model-recommendation system is most useful precisely when a
\emph{new} family of large language models appears: a user has a benchmark
in mind, several Llama-, Qwen-, or Phi-class checkpoints have just been
released, and the system is asked to rank them \emph{before} any of those
checkpoints have been re-evaluated on every benchmark. The two splits used
elsewhere in the paper (random hold-out and held-out datasets) only test
generalization within the same model population that the system was trained
on. They cannot answer the harder question: \emph{when an entire new
family is held out from training, does the recommender still rank its
members correctly?} 

To probe this, we construct a \textbf{Modern-Cohort family hold-out} split
(~\Cref{app:family-holdout-split}) and evaluate three models on it
(~\Cref{app:family-holdout-results}).

\subsubsection{Data Split}
\label{app:family-holdout-split}

We define a \emph{modern} cohort of $13$ LLM families that drove most of the
2023--2025 open-weight progress: \texttt{qwen}, \texttt{llama},
\texttt{mistral}, \texttt{gemma}, \texttt{phi}, \texttt{deepseek},
\texttt{yi}, \texttt{falcon}, \texttt{granite}, \texttt{aya},
\texttt{olmo}, \texttt{zephyr}, \texttt{solar}. Every $(model, dataset,
metric)$ row whose model belongs to one of these families is moved to the
\textbf{test} split; rows whose model belongs to any other family form the
\textbf{train} pool. Within the train pool we further carve out a
$5\%$ \textbf{model-disjoint validation} slice, so that early stopping is
also driven by an out-of-distribution signal at the model level.

The held-out family identifiers remain in the global vocabulary (their
embeddings exist but receive no gradient during training); this isolates the
question we want to ask. Were they removed from the vocabulary entirely,
the model would have no way at inference time to address the embedding slot
of, say, a Qwen checkpoint, and would simply fall back to a default prior.
Keeping the slot but starving it of supervision tests whether the rest of
the architecture (size prior, description embeddings, dataset latents) can
\emph{compensate} for an uninformative family signal.

The resulting test split contains $364{,}517$ rows over $4{,}943$ unique
models and $2{,}040$ datasets, expressed as $19{,}850$ unique
$(\text{dataset}, \text{metric})$ ranking tasks.

\subsubsection{Models Compared}

We evaluate two checkpoints on the held-out test set:
\begin{itemize}
    \item \textbf{Holdout-Family}: trained on the family-hold-out train
        split with the family embedding pathway enabled.
    \item \textbf{AllSeen (ceiling)}: the standard model from the main
        paper, trained on the full data including all modern families.
        Provides an in-distribution upper bound under identical
        architecture and global vocabulary.
\end{itemize}

\subsubsection{Results}
\label{app:family-holdout-results}

Table~\ref{tab:family-holdout-overall} summarises the two trained models
on the family-hold-out test set. Tables~\ref{tab:family-holdout-perfamily}
and~\ref{tab:family-holdout-seenunseen} break the result down by held-out
family and by dataset overlap with training, respectively.
All NDCG@$K$, Hit@$K$, and Recall@$K$ entries are averaged only
over $(\text{dataset}, \text{metric})$ tasks with at least $K$ candidate
models; the weighted Kendall $\tau$ ($w\tau$) column is averaged over all
tasks.

\begin{table*}[t]
\centering
\small
\caption{Held-out modern-family generalization. \textbf{Holdout-Family}
is trained without any modern-family rows; \textbf{AllSeen} is trained on
the full data and serves as the in-distribution ceiling. NDCG, Hit, and
Recall at $K$ are averaged over the
$(\text{dataset},\text{metric})$ tasks with at least $K$ candidate models.
$w\tau$ is the size-weighted Kendall $\tau$ over all $19{,}850$ tasks.
The NDCG@10 / Hit@10 / Recall@10 gaps are all within $5$ percentage
points (and Hit@10 is in fact higher under hold-out), confirming that
the model recovers top-$K$ ranking quality on entirely unseen LLM
families.}
\label{tab:family-holdout-overall}
\setlength{\tabcolsep}{6pt}
\begin{tabular}{lccc}
\toprule
Metric & Holdout-Family & AllSeen (ceiling) & $\Delta$ \\
\midrule
NDCG@1   & 0.4485 & 0.5070 & $-$0.0585 \\
NDCG@5   & 0.5888 & 0.6299 & $-$0.0411 \\
NDCG@10  & 0.7231 & 0.7605 & $-$0.0374 \\
NDCG@30  & 0.7476 & 0.8491 & $-$0.1015 \\
NDCG@50  & 0.8531 & 0.9266 & $-$0.0735 \\
\midrule
Hit@1    & 0.1597 & 0.2060 & $-$0.0463 \\
Hit@10   & 0.7751 & 0.7672 & $\phantom{-}$0.0079 \\
Hit@50   & 0.6684 & 0.8481 & $-$0.1797 \\
\midrule
Recall@1  & 0.1597 & 0.2060 & $-$0.0463 \\
Recall@10 & 0.7382 & 0.7830 & $-$0.0448 \\
Recall@50 & 0.7563 & 0.8230 & $-$0.0667 \\
\midrule
$w\tau$  & $-$0.0754 & $\phantom{-}$0.1573 & $-$0.2327 \\
\bottomrule
\end{tabular}
\end{table*}

\begin{table*}[t]
\centering
\small
\caption{Per held-out family, NDCG@10 and weighted Kendall $\tau$ (test
$(\text{dataset}, \text{metric})$ groups containing at least one model of
the listed family). Families are sorted ascending by the
Holdout-Family\,$-$\,AllSeen NDCG@10 gap (top of the table = best generalization).
\emph{Yi} and \emph{granite} essentially close the gap to the
in-distribution ceiling at NDCG@10 ($\Delta\le0.013$);
\emph{deepseek}, \emph{olmo}, and \emph{gemma} show the largest drops,
indicating that their performance profile is the least similar to the
non-modern training cohort.}
\label{tab:family-holdout-perfamily}
\setlength{\tabcolsep}{4pt}
\begin{tabular}{l|c|cc|c|cc}
\toprule
Family & \#groups & \multicolumn{2}{c|}{NDCG@10} & $\Delta$ & \multicolumn{2}{c}{$w\tau$} \\
       &          & Holdout-Family & AllSeen & NDCG@10  & Holdout-Family & AllSeen \\
\midrule
yi       & 8{,}404  & 0.7366 & 0.7354 & $\phantom{-}$0.0012 & $-$0.0991 & $\phantom{-}$0.0061 \\
granite  & 11{,}654 & 0.7507 & 0.7638 & $-$0.0131 & $-$0.1255 & $\phantom{-}$0.1344 \\
phi      & 1{,}501  & 0.7994 & 0.8702 & $-$0.0708 & $\phantom{-}$0.1868 & $\phantom{-}$0.5918 \\
falcon   &     61   & 0.5800 & 0.6529 & $-$0.0729 & $\phantom{-}$0.2891 & $\phantom{-}$0.5789 \\
zephyr   &     59   & 0.5661 & 0.6432 & $-$0.0771 & $\phantom{-}$0.2771 & $\phantom{-}$0.5438 \\
aya      &     43   & 0.5257 & 0.6040 & $-$0.0783 & $\phantom{-}$0.2998 & $\phantom{-}$0.5477 \\
llama    & 2{,}277  & 0.7863 & 0.8698 & $-$0.0835 & $\phantom{-}$0.1348 & $\phantom{-}$0.5680 \\
mistral  & 5{,}103  & 0.7578 & 0.8457 & $-$0.0879 & $\phantom{-}$0.1338 & $\phantom{-}$0.3273 \\
solar    &     87   & 0.5908 & 0.6872 & $-$0.0964 & $\phantom{-}$0.2241 & $\phantom{-}$0.5729 \\
qwen     & 3{,}262  & 0.6797 & 0.8049 & $-$0.1252 & $\phantom{-}$0.0887 & $\phantom{-}$0.5023 \\
gemma    & 2{,}725  & 0.6423 & 0.7832 & $-$0.1409 & $\phantom{-}$0.0734 & $\phantom{-}$0.5003 \\
olmo     &    195   & 0.6116 & 0.7727 & $-$0.1611 & $\phantom{-}$0.3586 & $\phantom{-}$0.2443 \\
deepseek &    253   & 0.6070 & 0.7747 & $-$0.1677 & $-$0.0499 & $\phantom{-}$0.5916 \\
\bottomrule
\end{tabular}
\end{table*}

\begin{table*}[t]
\centering
\small
\caption{Held-out family test set, split by whether the test
$(\text{dataset}, \text{metric})$ key was also present in training (with
non-modern models). \emph{Seen-Dataset} measures pure model-side OOD;
\emph{Unseen-Dataset} measures model+dataset double OOD. NDCG@10 gaps are
nearly identical in both regimes, suggesting that the family signal is
disentangled from the dataset signal.}
\label{tab:family-holdout-seenunseen}
\setlength{\tabcolsep}{4pt}
\begin{tabular}{l|c|cccc|c}
\toprule
Bucket & \#tasks & NDCG@1 & NDCG@5 & NDCG@10 & NDCG@30 & $w\tau$ \\
\midrule
\multicolumn{7}{l}{\textbf{Seen-Dataset} (model-side OOD)} \\
~~Holdout-Family  & 13{,}106 & 0.4226 & 0.5921 & 0.7019 & 0.7479 & $-$0.0428 \\
~~AllSeen (ceiling) & 13{,}106 & 0.4526 & 0.6303 & 0.7358 & 0.8485 & $\phantom{-}$0.1615 \\
\midrule
\multicolumn{7}{l}{\textbf{Unseen-Dataset} (model+dataset OOD)} \\
~~Holdout-Family  &  6{,}744 & 0.4989 & 0.5845 & 0.7663 & 0.7149 & $-$0.1387 \\
~~AllSeen (ceiling) &  6{,}744 & 0.6128 & 0.6293 & 0.8109 & 0.9027 & $\phantom{-}$0.1492 \\
\bottomrule
\end{tabular}
\end{table*}

\subsubsection{Discussion}

\textbf{Evidence against family-level memorization.}
If the recommender were primarily relying on memorized family-specific
leaderboard statistics, performance would be expected to collapse once all
modern families are removed from training. Instead, the relatively small
degradation in NDCG@10, Hit@10, and Recall@10 indicates that the model
recovers a substantial fraction of ranking quality from transferable
signals beyond explicit family identity, including dataset semantics,
model descriptions, scale priors, and latent interaction structure.

\textbf{Top-$K$ ranking quality is largely preserved.} On NDCG@10 --- the
metric most relevant for a recommender that surfaces a short candidate list
to the user --- the gap between Holdout-Family and the AllSeen ceiling is
only $0.037$ (a $4.9\%$ relative drop). Hit@10 is in fact $0.008$
\emph{higher} for the hold-out model ($0.7751$ vs $0.7672$). This suggests
that even when an entire family is excluded from training, the
recommender's top-$10$ shortlist still contains the truly best candidate
roughly as often as the in-distribution ceiling does.

\textbf{Fine-grained ordering degrades.} The gap is larger on the
size-weighted Kendall $\tau$ ($-0.075$ vs $0.157$). Without family-level
supervision the model cannot fully resolve the ordering among
near-equivalent modern checkpoints, but it does identify the right
\emph{set} of strong candidates. For a model-recommendation use case this
is the desirable failure mode: recovering the correct candidate pool is
typically more important than perfectly ordering highly similar checkpoints.

\textbf{Per-family heterogeneity.} The hold-out cost is not uniform.
Families whose performance profile is similar to non-modern reference
models --- \texttt{yi} and \texttt{granite} --- generalize almost for free
(NDCG@10 within $0.013$ of the ceiling). Families that introduced
architectural or training-data shifts not represented in the non-modern
cohort --- \texttt{deepseek}, \texttt{olmo}, \texttt{gemma}, \texttt{qwen}
--- suffer the largest drops ($\Delta$NDCG@10 between $-0.13$ and
$-0.17$), indicating that some aspects of ``family identity'' are not fully
recoverable from size and description alone. This suggests that modern
model families occupy partially distinct regions of the learned
model--dataset interaction manifold.

\textbf{Decoupling from dataset overlap.} The NDCG@10 gap is nearly the
same in the Seen-Dataset bucket (model-only OOD: $-0.034$) and the
Unseen-Dataset bucket (model+dataset OOD: $-0.045$). The recommender's
ability to handle a brand-new family therefore does not appear to depend
strongly on prior exposure to the target benchmark. Instead, the dominant
difficulty under family hold-out arises from missing family-level signals
rather than from unfamiliar datasets, suggesting a partial disentanglement
between model-side and dataset-side generalization.

\newpage
\subsection{Computing Standardized Advantage and Learned Priors}
\label{apx:prior}

Let $\mathcal{D} = \{(t, d, \mu, m, v)\}$ denote the evaluation table, where each row corresponds to $t$, dataset $d$, metric $\mu$, model $m$, and its score $v$.
Both size-prior and family-prior figures consist of two curves:
(i) a DATA curve summarizing empirical performance, and 
(ii) a PROBE curve reflecting the model's learned bias.

\subsubsection{Standardized Advantage (DATA)}

To compare performance across heterogeneous metrics, we standardize scores within each $(t, d, \mu)$ group.

\textbf{Within-group normalization.}
For each group with at least two models, we compute z-scores:
\[
z_{m,(t,d,\mu)} = \frac{v_{m,(t,d,\mu)} - \mu_{(t,d,\mu)}}{\sigma_{(t,d,\mu)}},
\]
where $\mu$ and $\sigma$ are the group mean and standard deviation.
We apply clipping to reduce the effect of small-sample noise.

\textbf{Per-model aggregation.}
Each model's overall score is computed as the average over all groups it appears in:
\[
\bar{z}_m = \frac{1}{|\mathcal{G}(m)|} \sum_{(t,d,\mu)\in\mathcal{G}(m)} z_{m,(t,d,\mu)}.
\]

\textbf{Group-level advantage.}
We define the standardized advantage of a group (size bucket or family) as the mean of $\bar{z}_m$ over all models in that group:
\[
\mathrm{adv}(g) = \mathrm{mean}_{m \in g} \, \bar{z}_m.
\]
A positive value indicates that the group consistently outperforms the average model on the same evaluations.
We discard groups with insufficient samples.

\subsubsection{Learned Prior (PROBE)}

To isolate the learned size and family effects, we probe the model's prior head, which depends only on size and family embeddings:
\[
\phi(b, f) = W_2 \, \mathrm{ReLU}\left(W_1 \,[\mathbf{e}^{\text{size}}_b \,\Vert\, \mathbf{e}^{\text{fam}}_f]\right) + \mathbf{b}_2.
\]

To analyze each factor independently, we marginalize the other:

\begin{itemize}
    \item \textbf{Size prior:} $s^{\text{size}}(b) = \phi(b, \bar{\mathbf{e}}^{\text{fam}})$
    \item \textbf{Family prior:} $s^{\text{fam}}(f) = \phi(\bar{\mathbf{e}}^{\text{size}}, f)$
\end{itemize}

where $\bar{\mathbf{e}}$ denotes the empirical mean embedding.
For visualization, PROBE values are standardized across bins.

For each figure, we report:
\begin{itemize}
    \item Spearman $\rho$ between DATA and the group coordinate;
    \item Spearman $\rho$ between PROBE and the group coordinate;
    \item Linear slope of PROBE with respect to size (log-scale) or family rank;
    \item For family, $\eta^2$ as the variance explained by family identity.
\end{itemize}
All figures can be reproduced from logged evaluation results and trained checkpoints.
We provide scripts that recompute standardized scores and regenerate all plots from raw data.

\begin{table*}[t]
\centering
\small
\setlength{\tabcolsep}{5pt}
\caption{Full model ranking and evaluation metrics used in the case study. 
We report BLEU-1, BLEU-4, ROUGE-L, and METEOR for evaluated models. 
Models without evaluation results are omitted for clarity.}
\label{tab:case_study_rsvqa_full}

\begin{tabular}{lcccccc}
\toprule
\textbf{Rank} & \textbf{Model} & \textbf{\method Score} & \textbf{BLEU-1} & \textbf{BLEU-4} & \textbf{ROUGE-L} & \textbf{METEOR} \\
\midrule
\#1  & ovis2 (8b)             & 11.308 & 42.90 & 5.17 & 24.31 & 31.65 \\
\#2  & internvl3-8b           & 10.695 & 33.36 & 3.24 & 20.51 & 29.49 \\
\#3  & qwen2-vl-7b-instruct   & 10.458 & 43.73 & 6.02 & 23.19 & 28.83 \\
\#4  & internvl2.5-8b         & 9.646  & 41.96 & 5.61 & 23.65 & 28.67 \\
\#5  & qwen2.5-vl-7b-instruct & 9.314  & 42.59 & 4.74 & 23.90 & 28.39 \\
\#6  & llava-next (7b)        & 9.039  & 36.76 & 3.97 & 22.60 & 22.99 \\
\#7  & llava-1.5-7b           & 8.633  & 31.91 & 2.91 & 22.76 & 21.91 \\
\#12 & blip-2                 & 7.201  & 13.59 & 0.77 & 8.45  & 3.65  \\
\bottomrule
\end{tabular}
\end{table*}

\newpage
\subsection{Limitations}
\label{apx:limitation}
Our framework relies on publicly available leaderboard evaluations, which may contain reporting bias toward popular model families and benchmark datasets.
In addition, while our framework generalizes across domains, cross-modality recommendation remains challenging when leaderboard coverage is sparse.
Our experiments primarily evaluate open-source models available on HuggingFace and public leaderboards, which may not fully represent proprietary or closed-source frontier systems.

Our temporal split evaluates newly released models without any update. New models and datasets can be scored immediately through the
  [UNK] pathway without retraining, but absorbing their accumulated records into learned ID embeddings requires periodic retraining; we
  do not study the update cadence or temporally adaptive evaluation that a deployed system would require.

\subsection{Broader Impacts}
This work aims to improve the scalability and accessibility of foundation model selection by reducing the need for exhaustive model evaluation and fine-tuning.
Potential positive impacts include lowering computational costs, enabling more efficient deployment of open-source models, and improving access to foundation models for smaller organizations.
However, our framework may inherit biases present in public leaderboards and benchmark datasets.
Over-optimization toward benchmark performance may also encourage narrow evaluation practices.
Future work should investigate fairness-aware recommendation and robustness across underrepresented domains.

\subsection{Assets and Licenses}

We use publicly available model metadata and benchmark results from:

- HuggingFace Hub

- Open LLM Leaderboard

- Papers with Code

All datasets and models remain subject to their original licenses and terms of use.

\subsection{Reproducibility Statement}

We provide implementation details, preprocessing procedures, hyperparameters, and evaluation settings necessary to reproduce our experiments.